\documentclass[11pt]{article}
\usepackage[top=0.9in,bottom=1.1in,left=1.05in,right=1.05in]{geometry}
\usepackage{amsmath,amssymb}
\usepackage{amsthm}
\usepackage{threeparttable}
\usepackage{latexsym}
\usepackage{mathrsfs,dsfont}
\usepackage{cancel}
\usepackage{comment}
\usepackage{float}
\usepackage{graphicx}
\usepackage{epstopdf}
\usepackage{color}
\usepackage{enumerate}
\usepackage{algorithm,algorithmic}
\usepackage{mathtools}
\mathtoolsset{showonlyrefs}
\DeclareGraphicsExtensions{.eps}
\usepackage{tabu}
\usepackage{subfigure}
\usepackage{hyperref}
\usepackage[font=small,labelfont=bf]{caption}
\usepackage{bm}

\numberwithin{equation}{section}

\newcommand{\XX}{\mathcal{X}}

\newcommand{\LL}{\mathcal{L}}
\newcommand{\FF}{\mathcal{F}}

\newcommand{\CC}{\mathcal{C}}

\newcommand{\bd}[1]{\boldsymbol{#1}}

\newcommand{\mc}[1]{\mathcal{#1}}
\newcommand{\mk}[1]{\mathfrak{#1}}

\newcommand{\MCC}{\mathcal{C}}

\newcommand{\MCF}{\mathcal{F}}

\newcommand{\EE}{\mathbb{E}}

\newcommand{\RR}{\mathbb{R}}

\newcommand{\eps}{\epsilon}

\newcommand{\abs}[1]{\left|#1\right|}

\newcommand{\ud}{\,\mathrm{d}}

\newcommand{\half}{\frac{1}{2}}

\newcommand{\dl}{\text{dl}}

\begin{document}

\title{\vspace{-50pt} Deep Learning for Ranking Response Surfaces with Applications to Optimal Stopping Problems}
\author{Ruimeng Hu\thanks{Department of Statistics, Columbia University, New York, NY 10027-4690, {\em rh2937@columbia.edu}. }
        }
\date{\today}
\maketitle

\begin{abstract}

In this paper, we propose deep learning algorithms for ranking response surfaces, with applications to optimal stopping problems in financial mathematics. The problem of ranking response surfaces is motivated by estimating optimal feedback policy maps in stochastic control problems, aiming to efficiently find the index associated to the minimal response across the entire continuous input space $\XX \subseteq \RR^d$.  By considering points in $\XX$ as pixels and indices of the minimal surfaces as labels, we recast the problem as an image segmentation problem, which assigns a label to every pixel in an image such that pixels with the same label share certain characteristics. This provides an alternative method for efficiently solving the problem instead of using sequential design in our previous work [R. Hu and M. Ludkovski, SIAM/ASA Journal on Uncertainty Quantification, 5 (2017), 212--239]. 

Deep learning algorithms are scalable, parallel and model-free, {\it i.e.}, no parametric assumptions needed on the response surfaces. Considering ranking response surfaces as image segmentation allows one to use a broad class of deep neural networks (NNs), {\it e.g.}, feed-forward NNs, UNet, SegNet, DeconvNet, which have been widely applied and numerically proved to possess good performance in the field. We also systematically study the dependence of deep learning algorithms on the input data generated on uniform grids or by sequential design sampling, and observe that the performance of deep learning is {\it not} sensitive to the noise and locations (close to/away from boundaries) of training data. We present a few examples including synthetic ones and the Bermudan option pricing problem to show the efficiency and accuracy of this method. We also simulate a ten-dimensional example to show the robustness, while non-learning algorithms in general have difficulties in such a high dimension.

\end{abstract}


\section{Introduction}\label{sec_intro} 

We start by introducing the general setup of ranking surface response problems, and then describe the connection to stochastic control problems and related literature. For the purpose of consistency, we shall use the same notions as in our previous work \cite{HuLudkovski17} and refer the interested readers to it for more details of the problem background.

Denote $\mu_\ell : \RR^d \supseteq \XX \to \RR$, $\ell \in \mk{L} \equiv \{1,2, \ldots, L\}$ as smooth functions over a subset $\XX$ of $\RR^d$. The surface ranking problem consists in assigning the index of the minimal surface  to every input $x$ in the entire (usually continuous) space $\XX$, namely, in finding the classifier
\begin{equation}\label{def_cal}
\mathcal{C}(x) := \arg\min_\ell \left\{\mu_\ell(x)\right\} \in \mk{L}, \quad \forall x \in \XX \subseteq \RR^d.
\end{equation}
The functions $\mu_\ell$ are {\it a priori} unknown but can be noisily sampled, {\it i.e.}, for any $x \in \XX, \ell \in \mk{L}$, one can access $\mu_\ell(x)$ through its stochastic sampler $Y_\ell(x)$:
\begin{equation}\label{def_Y}
Y_\ell(x) = \mu_\ell(x) + \eps_\ell(x), \ \ell \in \mk{L},
\end{equation}
where $\eps_\ell$'s are independent random variables with mean zero and variance $\sigma_\ell^2(x)$. 
In other words, one can sample by Monte Carlo the $L$ smooth hyper-surfaces on $\XX$.  

We are interested in accurately estimating $\CC(x)$ using deep learning algorithms.  Let  $\CC_{dl}(x)$ be the classifier produced by deep learning algorithms, and to study the performance of different network architectures and experiments designs, we evaluate the difference by the following loss metric:
\begin{equation}\label{def_loss}
\LL(\CC, \CC_{dl}) = \int_{\XX}  \mathds{1}_{\{\CC(x) \neq \CC_{dl}(x)\}}\lambda(\mathrm{d}x),
\end{equation}
where $\lambda(\mathrm{d}x) \in \mathcal{P}(\XX)$ is a probability  measure on $\XX$ specifying the relative importance of ranking different regions. Note that when $\lambda(\mathrm{d}x) = \mathcal{U}(\XX)$ is the uniform measure, $\LL$ gives the mis-labeled percentage. This is essentially the \emph{accuracy} metric in deep learning, often appearing as the first metric one uses to judge the performance of a neural network model/architecture.

To interpret \eqref{def_Y} in the context of dynamic programming, one can think of $x$ as system states, $\ell$ as indices of various actions available to the controller, $\mu_\ell(\cdot)$ as the expected costs-to-go and $\eps_\ell(\cdot)$ as simulation noises arising from pathwise evaluation of the underlying stochastic system and corresponding costs. In what follows, we will clarify this relation.

\medskip
\noindent\textbf{Motivation.} We consider the following stochastic control problems, and let $X_{(\cdot)} \equiv X_{(\cdot)}^u$ be a discrete-time stochastic state process controlled by a Markovian strategy $u_{0:T-1} \in \mk{L}^T$, following 
\begin{equation}
X_{t+1}^u = F(X_t, u_t, \varepsilon_{t+1}), \quad t \in \{0,1, \cdots, T-1\},
\end{equation}
for some map $F: \XX \times \mk{L} \times \RR \to \XX$, and some centered independent noise source  $\varepsilon_{t+1}$. Note that we use a single subscript $t$ to represent the value of the process at time $t$, and use  $t:T$ to emphasis the whole $\{\MCF_t\}$-adapted process from $t$ to $T$, $\MCF_t = \sigma(\varepsilon_1, \ldots \varepsilon_t)$. In general, the cost to be minimized is of the form
\begin{equation}\label{def_cost}
 c(0,u_{0:T}) = \sum_{t = 0}^T g(t, X_t^u, u_t), 
\end{equation}
where $g(t,\cdot, u_t)$ represents the running cost at stage $t$ using the strategy $u_t$. By defining the value function 
\begin{equation}\label{def_V}
V(t,x) := \inf_{u_{t:T} \in \mk{L}^{T-t+1}} \EE_{t,x}[c(t, u_{t:T})],
\end{equation}
with $E_{t,x}[\cdot] = E[\cdot \vert X_t = x]$ denoting the expectation given condition $x$, the minimized cost is represented by $V(0,x)$ and the corresponding optimal strategy is given by the minimizer $u^\ast_{0:T}$. Using dynamic programming ({\it e.g.} \cite{bjork2009arbitrage}), $V(t,x)$ satisfies:
\begin{equation}\label{eq_V}
V(t,x) = \inf_{u \in \mk{L}} \{g(t,x,u) + \EE_{t,x}[V(t+1,X_{t+1}^u)]\}.
\end{equation}

By introducing the term $\mu_u(t,x)$, called Q-value in \cite{HuLudkovski17}:
\begin{equation}\label{def_muu}
\mu_u(t,x) = g(t,x,u) + \EE_{t,x}[V(t+1, X_{t+1}^u)], \quad  \forall u \in \mk{L},
\end{equation}
the connection to problem \eqref{def_cal}--\eqref{def_Y} now becomes clear. For each $u \in \mk{L}$, the Q-value $\mu_u(t,x)$, representing the expected cost-to-go corresponding to action $u$, is a response surface in \eqref{def_Y}. For fixed $t$, finding the optimal strategy map $x \mapsto u^\ast(t,x)$ is equivalent to identifying the classifier in  \eqref{def_cal}, as $V(t,x) = \inf_{u \in \mk{L}}\left\{\mu_u(t,x)\right\}$. Then, this stochastic control problem $V(0,x)$ can be solved by backwardly identifying $u^\ast(s,\cdot)$, namely, by solving $T$ ranking problems of the form \eqref{def_cal} from stage  $T$ to 1. More precisely, assuming strategy maps after time $t$, denoted by $\left\{\hat u(s,\cdot)\right\}_{s=t+1}^T$, are already generated, then $\hat u(t, \cdot)$ is determined by ranking \eqref{def_muu} across different $u$, where $\EE_{t,x}[V(t+1, X_{t+1}^u)]$ is estimated by averaging the cost \eqref{def_cost} along trajectories $X_{(t+1):T}$ that follow the strategy $\left\{\hat u(s,\cdot)\right\}_{s=t+1}^T$. 

In principle, this approach is applicable to any stochastic control problem, including continuous-time framework with a continuum of strategies, since both time and strategy space can be approximated by discretization. However, it is especially attractive when the number of actions is finite and small. For instance, in optimal stopping problems \cite{GL13},
the action space has only two elements $\mk{L} = $\{stop, continue\} and the immediate reward $\mu_{stop}$ is usually obtained at no cost, leading to only the continuation value $\mu_{cont.}(t,x)$ to be simulated. A canonical example in this case is pricing Bermudan-type options \cite{ludkovski2015kriging}. Applications that need to evaluate multiple surfaces $\mu_\ell$ arise from pricing swing options \cite{MeinshausenHambly04}, decision makings in energy markets \cite{AidLangrene12,Secomandi11} ({\it i.e.}, deciding whether to expand the productivity, to explore new energy resources or do nothing), epidemic management \cite{LudkovskiLin14,LN10,LN11wsc,MerlGramacy09}, to name a few. 

\medskip
\noindent\textbf{Main approach and contribution.}
Our main contribution is to propose an alternative strategy to solve \eqref{def_cal} by deep learning algorithms. The key idea is to build up a neural network (NN) and let it learn to solve the problem of interest by itself via simulated data. This learning process, which is called the training of an NN, can be time-consuming. However, once this is accurately done, it will be computationally efficient to solve a problem of the same type. In our case, this is to say, once we use certain simulated data $Y_\ell(x)$ and its labels to train the neural network and obtain the desired accuracy, the predicted classifier $\CC_{dl}$ of a new location $x$ will be instantaneous and accurate by mainly operations of matrix-vector multiplication. For this reason,  NN is a desired approach to solve \eqref{def_cal}.

The problem of ranking response surfaces is equivalent to partitioning the entire input space $\XX$ into parts distinguished by labels (indices of the minimal surface). We observe that, if one treats $\XX$ as an image, then the labeling function $\mathcal{C}$ essentially divides the image into disjoint parts. This means that one can phrase the problem as image segmentation, where deep learning has been successful and become a primary and powerful tool in modern machine learning community \cite{LeCunBengioHinton15,ShelhamerLongDarrell15,RonnebergerFischerBrox15}. In the meantime, mathematical theory on deep neural networks (NNs), \emph{e.g.}, whether results produced by NN converge to the ground truth as the number of neurons/layers tends to infinity, has also been developed by analyzing the corresponding mean-field optimal control problem \cite{EHanLi19}. Compared to our previous work \cite{HuLudkovski17} where response surfaces are modeled by Gaussian process, the advantage of deep learning algorithms is that it is model-free, that is, they make the predicted labels $\hat \CC$ no more depend on specific parameterizations of $\mu_\ell$. Moreover, through numerical studies, we find that NN algorithms also have the following advantages:
\begin{itemize}
	\item It is insensitive to sampling locations. The loss $\LL$ are comparable when  $C_{dl}$ are produced using uniform samples versus sequentially designed samples \cite{HuLudkovski17} over $\XX \times \mk{L}$. Then regarding implementation complexity and the ability of parallelism, uniform sampling is more preferable.
	\item It can auto-detect wrong inputs. Since $\mu_\ell$ is only accessible by its stochastic sampler $Y_\ell$, the training input labels are $\arg\min_{\ell \in \mk{L}} Y_\ell$, which certainly contain wrong labels especially around the boundaries. The NN will try not to learn those labels correctly, and can automatically ignore those inputs.
\end{itemize}

\medskip
\noindent\textbf{Related Literature.}
Mathematically, one can view \eqref{def_cal} as a partition over the input $\XX = \cup_{i=1}^L \CC_i$:
\begin{equation}
\CC_i := \{x \in \XX, \CC(x) = i\}, i \in \mk{L}.
\end{equation}
The problem is related to contour-finding of $\partial \CC_i$, which has been extensively studied by numerous sequential methods \cite{GL13,Picheny10,RanjanBingham08}. For each $x$, the goal of identifying the minimal response $\arg\min_\ell\mu_\ell(x)$ corresponds to multi-armed bandits (MAB) problems. Consider the surfaces $\mu_\ell(x)$ as $L$ arms' rewards of the bandit, then \eqref{def_cal} is equivalent to exploring the extreme bandit \cite{BubeckMunos11,BubeckMunos11X,GabillonBubeck11,GrunewalderAudibert10}. 
Statistically, for a tractable approximation of $\mu_\ell$, various models have been proposed, including Gaussian process (GP) \cite{HuLudkovski17}, BART \cite{chip:geor:mccu:2010}, Dynamic trees \cite{GTP-trees11}, treed GPs  \cite{tgpPackage}, local GPs \cite{gramacy:apley:2013}, particle based Gaussian process \cite{GramacyPolson11}, GPs with Student-$t$ noise and $t$-processes \cite{lyu2018evaluating}.

Let us mention two recent works that are related to our paper. In our previous work \cite{HuLudkovski17}, the problem \eqref{def_cal} was tackled under the GP modeling for $\mu_\ell$ with a different loss metric:
\begin{equation}\label{eq:loss}
\LL(\hat{\CC}, \CC) := \int_\XX \left\{ \mu_{\hat{\CC}(x)}(x) -\mu_{\CC(x)}(x) \right\} \; \lambda(\mathrm{d} x).
\end{equation}
This is a blended criterion between marginal estimation of $\mu_\ell$ and classification. The loss is proportional to the difference between the true minimal surface and the estimated minimal one, which tolerates estimation errors of $\mu_\ell$ as long as the minimal response does not change. While in this paper, we make no model assumption on $\mu_\ell$ and treat \eqref{def_cal} as a pure classification/segmentation problem. 
In \cite{BeChJe:18}, Becker, Cheridito, and Jentzen directly address the optimal stopping problem using deep learning. They learn the optimal stopping rule via maximizing the stopped payoff along each Monte Carlo path via a feedforward NN with three fully connected layers. Compared to the results in \cite{BeChJe:18}, our work distinguishes for two reasons. On the one hand, our problem setup \eqref{def_cal} is more general, and optimal stopping problems is just an application of ranking response surfaces; on the other hand, the emphasis of our work is on the architecture of neural networks, {\it i.e.}, by recasting optimal stopping as the image segmentation problem, one is allowed to use a broader class of networks with delicate architecture designed for image segmentation in computer science literature (\emph{e.g.} convolutional neural networks \cite{ShelhamerLongDarrell15}, UNet \cite{RonnebergerFischerBrox15,HaRoMyYa:18}, SegNet \cite{BadrinarayananKendallCipolla17}), from which one can choose the best performance empirically. Note that, there is no existing result in literature that rigorously discusses which architecture produces the optimal performance of image segmentation, to our best knowledge.

\medskip 
\noindent\textbf{Organization of the paper.} The rest of the paper is organized as follows: In Section~\ref{sec:nn}, we introduce the design of network models and deep learning algorithms. In Section~\ref{sec:numerics}, we test the performance of deep learning by one-, two- and ten-dimensional examples of ranking response surfaces, and systematically study the dependence of deep learning algorithms on the quality of input data generated by uniform or by sequential design sampling. We apply the deep learning algorithms to Bermudan option pricing in Section~\ref{sec:Bermudan}, and make conclusive remarks in Section~\ref{sec:conclusion}.

\section{Neural networks and deep learning algorithms}\label{sec:nn}
Inspired by neurons in human brains, neural networks (NNs) are designed for computers to learn from observational data. Deep learning algorithms are techniques for accurate and efficient learning in neural networks. For interesting problems including image recognition, natural language processing, boundary detection,  image classification and segmentation, neural networks and deep learning currently provide the best solutions. In what follows, we give a brief introduction to basic concepts in neural networks and how it works. Section~\ref{sec:algorithm} is dedicated to the algorithms for our ranking problem \eqref{def_cal}.

\subsection{Preliminaries on deep learning} 
\begin{figure}[H]
	\centering \includegraphics[width=0.6\textwidth]{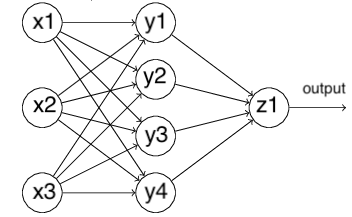}
	\caption{An illustration of a simple feedforward neural network. }\label{fig:ffnn}
\end{figure}

We start with some terminology. Basically, NNs are built up by {layers}. Each layer contains a number of {neurons}. Layers with different functions or neuron structure are called differently, including fully-connected layer, constitutional layer, pooling layer, recurrent layers, etc. Figure~\ref{fig:ffnn} below is a simple feed-forward NN with three fully-connected layers, where nodes represent neurons and arrows represent the information flow. As shown in the figure, information is constantly ``fed forward'' from one layer to the next. The first layer (leftmost column) is called the input layer, and the last layer (rightmost column) is called the output layer. Layers in between are called hidden layers, as they have no connection with the external world. In this case, there is only one hidden layer with four neurons. 

We now explain how NN learns from data. For fully-connected layers, every neuron has two kinds of parameters, the weights $w$ and the bias $b$.  An input $x$ goes through a neuron outputs $f(w \cdot x + b)$, where $f$ is an activation function. 
In the above illustrated NN,  $x_1$, $x_2$ and $x_3$ are the input of neural network. 
Nodes in the hidden layer take $\bd{x}=(x_1,x_2,x_3)$ as inputs and output $y_j = f(\bd{w}_j \cdot \bd{x} + {b}_j)$, $j=1,\cdots,4$, $\bd{w}_j=(w_{j,1},w_{j,2},w_{j,3})$. Then they are considered as inputs for the output layer, and $z_1=f(\bd{w}_z \cdot \bd{y} + {b_z})$.  Based on the training data set, {\it i.e.}, known pairs of input and desired output $z(x)$,  the {learning} of an NN is to find the optimal weights and biases, so that the output from the network well approximates $z$ for all training inputs $x$. Mathematically, this is done by minimizing some loss function, for instance, the mean square loss:
\begin{equation}\label{def_costfunctional}
c(w,b) = \frac{1}{2n}\sum_x \left\lVert z(x) - z\right\rVert^2,
\end{equation}
where $(w,b)$ denotes the collection of all weights and biases in the network and $n$ is the total number of training inputs. 
Depending on the depth and size of the NN, training can take hours. However, when this is done, that is, the optimal or near-optimal parameters are obtained, prediction from new input $x_0$ is efficient as it will be mostly matrix-vector multiplication. 

\medskip 
\noindent\textbf{Activation function.} Popular types are sigmoid, Tanh, ReLU, Leaky ReLu,  softmax,  etc. The activation functions are not necessarily the same from layer to layer, while for the output layer, some types generally work better than others. For instance, in binary classification problems, a common choice is the {sigmoid} function $f(x) = \frac{1}{1 + e^{-x}}$, as it maps real-valued inputs to $[0,1]$. For multiple-class classification, the {softmax} function, mapping $\RR^d$ to a probability simplex, is usually preferred. In both cases, outputs are interpreted as the probability of being in one of the categories.

\medskip
\noindent\textbf{Loss function.} Depending on the different goals, the form of loss function can be different, varying from mean squared error (MSE), mean absolute error, Kullback Leibler (KL) Divergence, $\ell_2$ norm, $\ell_1$ norm,  cross entropy, hinge to squared hinge.  The MSE loss is generally better suited to regression problems, while cross entropy and hinge are commonly used in classification problems. Besides, 
 $L_1$ or $L_2$ regularization are sometimes added to the loss function, to help to reduce overfitting. 

\medskip
\noindent\textbf{Optimizer.} Finding the optimal weights and biases in \eqref{def_costfunctional} is in general a high-dimensional optimization problem. This is so-called the training of NN, which is commonly done based on stochastic gradient descent method ({\it e.g}., Adam \cite{Adam, Adamcvg}, NADAM \cite{Dozat16}).

\subsection{Deep learning algorithms}\label{sec:algorithm}

In computer vision, images segmentation is the process of partitioning a digital image into multiple segments.  Each pixel in an image will be labeled for a class it belongs to. The training data consist of the RGB values of each pixel and its desired label. Considering ranking response surfaces as the segmentation of images, the ``image'' is then the entire input space $\XX$ while the class label is the index of the minimal surface. Each point in $\XX$ is treated as our ``pixels'', and the ``color'' of each ``pixel'' is then the coordinates of the location. With the concepts introduced above, we describe the deep learning algorithm with details in this subsection, which includes the input, output, and architecture of neural networks. 

\medskip
\noindent{\bf Input and output}. The design of the input and output layers in a network is often straightforward. Recall our problem \eqref{def_cal}, we aim at training an NN using noisily sampled data $Y_\ell(x)$ at some locations $x$, so that after training it can efficiently and accurately label each location $x$ in $\XX$ by the index of the minimal surface. The input to the network is $x^{1:J}\in\RR^{J\times d}$ where $J$ corresponds to the number of points and $d$ is the dimensionality of the problem. The desired output should take the form:
\begin{equation}
	\mathcal{C}_{\dl}=(p_{j\ell}),\quad j=1,\cdots,J,\; \ell=1,\cdots,L,
\end{equation}
where $L$ is the number of response surfaces, and $p_{j\ell}$ is the probability that the $\ell^{\text{th}}$ surface is the minimal at the $j^{\text{th}}$ point. This is usually achieved by implementing the softmax activation function for the output layer. 
For example, if one gets the following output
\begin{equation}\left(
\begin{matrix}
	0.1 & 0.2 & 0.7 \\
	0.8 & 0.1 & 0.1
\end{matrix}\right),
\end{equation}
then it means that the network believes the $3^{\text{rd}}$ surface is minimal with probability $70\%$ at the $1^{\text{st}}$ point, and that the $1^{\text{st}}$ surface is minimal with probability $80\%$ at the $2^{\text{nd}}$ point. The predicted labels for the corresponding points will be given by taking the row-wise argmax of the matrix, which produces a column vector in $\RR^J$. In the above example, the input contains two points $x^1$ and $x^2$, and the corresponding labels are $[3,1]^\dagger$. 

During the training stage, the network is told what the true labels should be for the inputs $x^{1:J}$, and it adjusts its belief according to this information via minimizing some loss function (cf. \eqref{def_costfunctional}). In the generalization stage, only locations in $\XX$ are given, and one uses the network output as the final result. In both stages, the accuracy will be evaluated by the percentage of correctly predicted labels against the ground truth, and results are called the \emph{training accuracy} and the \emph{generalization accuracy}. This is also the loss metric \eqref{def_loss} with uniform measure $\lambda(\mathrm{d}x) = \mathrm{d}x/\abs{\XX}$.

When training with noisy data, the ground truth is unknown and the true labels are up to our best knowledge.  That is,  the `` true'' label is produced by simulating $Y_\ell(x)$ for each $\ell \in \mk{L}$ and take $\arg\min_\ell Y_\ell(x)$. Of course, this leads to mis-labeling and affects both training and generalization accuracies.  In Section~\ref{sec:numerics}, we present numerical studies on synthetic examples with known ground truth and study the accuracies of deep learning algorithms.

\begin{figure}[H]
	\begin{tabular}{cc}
		\hspace{10em}\includegraphics[width=0.21\textwidth, height = 0.5\textheight]{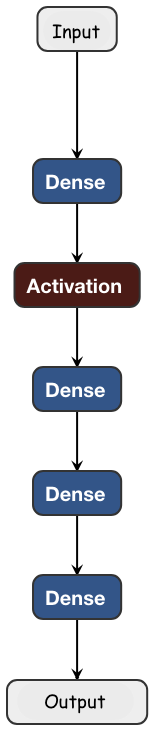} & 
		\hspace{10em}\includegraphics[width=0.25\textwidth, height = 0.5\textheight]{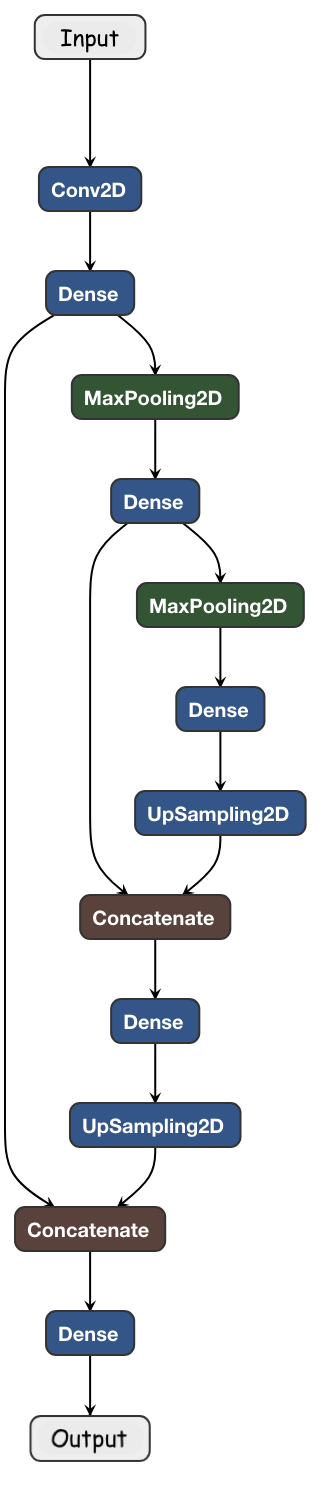}
	\end{tabular}
	\caption{Architectures of neural networks (NNs). Left: a feed-forward NN with 4 fully-connected hidden layers; Right: two-layered UNet. Here the dense block means a fully-connected layer. The purpose of adding an  activity regularizer is to reduce the generalization error, and may not always be needed. The ``MaxPooling2D'' is for downsampling, and ``Concatenate'' merges outputs from different dense blocks, which makes the architecture look like a ``U''-shape. }\label{fig:arch}
\end{figure}

\noindent{\bf Architecture}. Although the design of the input/output layers of an NN is usually straightforward, it can be quite sophisticated to find a good architecture for the hidden layers. In particular, it is unlikely to summarize the design strategies for the middle layers with a few simple rules, for instance,  how to trade off the number of hidden layers against the time required to train the network. 
The global topology of the network consists of blocks and the interactions between  them, which are in general described by the meta-architecture. Designing proper meta-architecture can improve the performance of networks, and in the context of ranking response surfaces, recasting the problem as image segmentation will allow us to use the meta-architecture of CNNs \cite{ShelhamerLongDarrell15}, UNet \cite{RonnebergerFischerBrox15,HaRoMyYa:18}, and SegNet \cite{BadrinarayananKendallCipolla17}, which has been considered and widely used as an efficient network structure for image segmentation. 

As there is no rigorous discussion on which architecture performs the best for image segmentation, we shall focus on two architectures in this paper: feed-forward NNs and UNet visualized
in Figure~\ref{fig:arch}. They are built using Keras \cite{Keras}, a high-level neural networks API.
Main blocks are fully-connected (shown as dense in Figure~\ref{fig:arch}) layers and  convolutional layers. The first (left) one is built up by dense blocks. We add an $\ell_1/\ell_2$ activity regularizer (shown as Activation) to help reduce the generalization error, which may not always be needed. ReLu is chosen as the activation function for hidden layers, while sigmoid or softmax is used for the output layer depending on the number of classifiers. In the second (right) architecture, we use a 2D convolutional layer (shown as Conv2D) with $3 \times 3$ kernel as the first hidden layer, which outputs 4D tensor. It expands the number of feature channels, which help to reduce the number of necessary feature maps leading to an improvement of computational efficiency \cite{HeZhReSu:15,SpDoBrRi:14}. ``MaxPooling2D'' is for downsampling, and ``Concatenate'' merges outputs from different dense blocks, with one before downsampling and one after upsampling so that their dimension is matched. 
The ``concatenate'' action makes the architecture have a ``U''-shape. Activation functions are chosen the same as in the first architecture. We remark that, such a UNet structure can be applied for problems with any dimensionality, by merely adjusting the dimensionality of convolutional and down/up sampling layers.

Let us also mention that the neural networks used in \cite{BeChJe:18} are a feed-forward NN with two fully-connected hidden layers, and what they actually make use of NN is its ability of approximating complex relations by compositions of simple functions (by stacking fully connected layers) and finding the (sub-)optimizer with its well-developed built-in stochastic gradient descent (SGD) solvers, whose convergence has been studied in literature ({\it e.g.,} \cite{Ho:91}). While by recasting the optimal stopping problems as image segmentation, one is allowed to use a broader class of neural networks with more delicate architecture ({\it e.g.} UNet), for which the convergence theory is still unclear.

\section{Numerical experiments}\label{sec:numerics}

In this section, we first analyze the performance of deep learning algorithms (both feed-forward NNs and UNet) by studying the one- and two-dimensional examples used in \cite{HuLudkovski17}, and systematically analyze the dependence of deep learning algorithms on the input data generated by uniform or by sequential design sampling. We also present a ten-dimensional example to show the robustness of deep learning algorithms, where non-learning algorithms in general have a difficulty in computational time. As a consistency, we shall use the same notations as in \cite[ Section 4]{HuLudkovski17}.

\subsection{One-dimensional example}
We consider the one-dimensional toy model used in \cite{HuLudkovski17}, originally from \cite[Section 4.4]{kmPackage-R}. Let $L = 2, \XX = [0, 1]$ in \eqref{def_cal}, and define the noisy responses $Y_1(x)$ and $Y_2(x)$ as 
\begin{align*}
Y_1(x) &= \mu_1(x) + \eps_1(x) \equiv \frac{5}{8}\left(\frac{\sin(10x)}{1+x} + 2x^3\cos(5x)+0.841\right) + \sigma_1(x)Z_1, \\
Y_2(x) &= \mu_2(x) + \eps_2(x) \equiv 0.5 + \sigma_2(x)Z_2,
\end{align*}
where $Z_\ell$ are independent standard Gaussians, with the noise strengths fixed at $\sigma_1(x) \equiv 0.2$ and $\sigma_2(x) \equiv 0.1$, homoscedastic in $x$ but heterogenous in $\ell=1,2$. We take the uniform weights $\lambda(\mathrm{d} x)=\ud x$ in the loss function on $\XX$, which is interpreted as the percentage of mis-labeled locations.

\begin{figure}[t]
	\centering\includegraphics[width=0.5\textwidth]{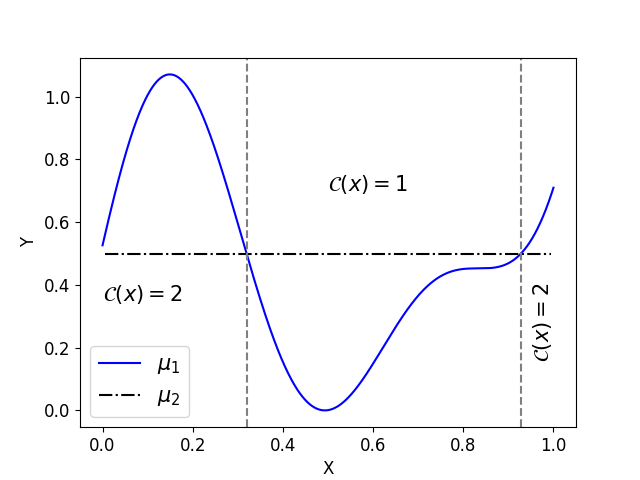}
	\caption{The true response surfaces $\mu_1$ and $\mu_2$ and the corresponding ranking classifier $\CC$ for the one-dimensional example. The entire input space $[0, 1]$ is divided into three parts, with ranking classifier equal to $1$ in the middle, and equal to $2$ otherwise.}\label{fig:1d_fun}
\end{figure}

Then the true ranking classifier $\CC(x)$ is computed as
\begin{equation}\label{def_label1d}
\CC(x) = \left\{ {\begin{array}{*{20}{l}}
	{2}& \quad \text{for } x \in [0,r_1] \cup [r_2,1], \\
	{1}& \quad \text{for } {r_1 < x < r_2, } \\
	\end{array}} \right.
\end{equation}
where $r_1 \approx 0.3193, r_2 \approx 0.9279.$ We visualize the true responses $\mu_1$ and $\mu_2$ and the corresponding ranking classifier $\CC$ in Figure~\ref{fig:1d_fun}.

We first investigate the performance of feed-forward NNs, and test them for the following four cases of training data generated on uniform grids or the points produced by sequential design. 
\begin{table}[H]
		\centering
		\caption{Summary of different design for $(x, z(x))$}\label{tbl:datadesign}
		\begin{tabu}{|l|c|c|}
			\hline
		Name of Method & Method to generate $x \in \XX$ & Method to generate labels \\ \tabucline[1pt]{1-3}
		UNIF & uniform grids & True label $\arg\min_\ell \mu_\ell(x)$ \\ \hline
		UNIF + NL & uniform grids & Noisy label $\arg\min_\ell y_\ell(x)$ \\ \hline
		SD & sequential design & True label $\arg\min_\ell \mu_\ell(x)$ \\ \hline
		SD + NL & sequential design & Noisy label $\arg\min_\ell y_\ell(x)$ \\ \hline
		\end{tabu}
	\end{table}
	
\noindent Let $M$ be the size of training data. For a comprehensive study, we conduct our experiments under different $M = 128, 256, 512$. Points generated by sequential design use ``Gap-SUR'' method developed in \cite{HuLudkovski17}, and are mainly concentrated near the boundaries $\partial \CC_i$, namely, around $r_1$ and $r_2$, as well as the ``fake'' boundary $x=0$, where the two lines are quite close but do not touch each other. Then labels are generated by taking the argmin of true surfaces $\mu_\ell$ or realizations $y_\ell$ of the noisy sampler $Y_\ell$ at those points $x^{1:M}$.

To focus on the performance of data designs in Table~\ref{tbl:datadesign}, we fix the network architecture to be a feed-forward NN with two fully-connected hidden layers. Number of neurons in each hidden layer is set at $M/8$. For this binary classification problem, the output layer contains one neuron and produces the probability of Surface $1$ being the minimum it believes, by using the sigmoid activation function. The network is trained for 1500 epochs with updating each gradient using $M/2$ data. One epoch is an iteration over the entire data. We remark that all above settings can be altered for a suitable trade-off between accuracy and efficiency. However, for a purpose of comparison, we fix them in the way we explained above. Meanwhile, we do adjust the learning rate and other parameters in the optimizer, in order to better train the network within 1500 epochs. 

In deep learning, two numbers are of most importance, the training accuracy and the generalization accuracy. The first one indicates how well the network learns from the training data set, calculated based on $M$ points, while the latter one measures the accuracy of the prediction for new locations $x \notin x^{1:M}$, calculated on a finer grid on $\XX$.  Also, note that the latter one is a discretized version of  $1 - \LL(\CC, \CC_{dl})$ where $\LL$ is the loss defined in \eqref{def_loss} with uniform measure. For these two reasons, we report accuracies instead of loss in Table~\ref{tbl:1d} for different computational budget $M = 128, 256, 512$ and different designs listed in Table~\ref{tbl:datadesign}.

\begin{table}[H]
	\centering
	\caption{Training accuracy versus generalization accuracy for the 1-D example with different computational budget $M$. The acronyms used are: UNIF = uniform grids on $\mathcal{X}$, SD = grids generated by Gap-SUR in \cite{HuLudkovski17}, NL = training with noisy label. \label{tbl:1d}}
	\begin{tabu}{|l|cc|cc|cc|}
		\hline
		{Method/Budget} & \multicolumn{2}{c|}{\textbf{M = 128}}   & \multicolumn{2}{c|}{\textbf{M = 256}}  & \multicolumn{2}{c|}{\textbf{M = 512}}    \\ 
		& Train. Acc. & Gen. Acc. & Train. Acc. & Gen. Acc. & Train. Acc. & Gen. Acc. \\ \tabucline[1pt]{1-7}
		UNIF & 99.9\%& 99.7\%& 99.9\%&99.9\%&99.9\%& 99.9\% \\
		UNIF + NL &81.25\%&98.5\%&79.3\%&98.8\%&81.0\%&99.5\%\\ \hline
		SD&99.5\%&98.3\%&96.1\%&98.9\%&98.1\%& 99.5\% \\
		SD + NL &64.1\%&97.3\%&57.2\%&92.3\%&58.2\%&94.2\%\\ \hline	
	\end{tabu}
\end{table}

We observe that, the training accuracy is higher than the generalization accuracy for NNs trained by clean data, while smaller than generalization accuracy for NNs trained by noisy data. This is because that, the generalization accuracy is tested on clean data, and the usage of noisy labels in training data set decreases the training ``accuracy'', evidenced by comparing UNIF to UNIF+NL. In fact, when there are errors in the training data set, the NN auto-detects these errors and avoid learning from them. By a careful examination, the misclassified locations in the training data set are mainly mis-labeled points due to the low signal-to-noise ratio, with a small number of points around the boundary, which are originally hard to learn. This can be understood in the sense that, although the training data with noisy labels do not contain $100\%$ accurate information, networks ``learn'' to minimize the effect of the wrong labels by not training them correctly. Secondly, by comparing UNIF+NL to SD+NL, we observe that the usage of SD further decreases the training accuracy. This is due to the fact that the input data of SD contain more errors, as points $x^{1:M}$ generated by SD are mostly concentrated around the boundaries $r_1 = 0.3193$, $r_2 = 0.9279$ and the fake boundary $x = 0$, where the signal-to-noise ratio is low, leading to a large number of mis-labeling points. Thirdly, we observe that there exists a threshold on the proportion of error in training data so that they can be auto-detect and will not influence the network's predicting accuracy. For instance, comparing SD with SD+NL at $M = 256$, the generalization accuracy significantly decreases. We interpret this phenomenon as the fact that there have been enough wrong labels which make the network believe they (the wrong labels) are the truth. 

Finally, we comment that, as increasing the budget for the simulation of training data set, the results are better in the UNIF case, which is consistent with common sense: the more data, the better the learning. While in the SD case, there is a turning point in both training and generalization accuracy, \emph{i.e.}, $64.1\% \to 57.2\% \to 58.2\%$, and $97.3\% \to 92.3\% \to 94.2\%$. This is because of the Gap-SUR criterion we use, where $x^{1:M}$ are sequentially selected by reducing stepwise uncertainty. When we adaptively grow $x^{1:M}$, the algorithm will mostly pick points around the boundaries first. The additional budget from $M = 128$ to $M = 256$ mostly goes to the boundary points, which increases the percentage of wrong labels, leading to a decrease in accuracy. Once the points there become saturated (the uncertainty reduction becomes very small), Gap-SUR favors locations that have large posterior variance, usually interior points of $\partial\CC_i$. Therefore, the additional 256 points in $M = 512$ case go to interior points more than the boundary ones, which increases the  accuracy.

\begin{figure}[H]
	\begin{tabular}{cc}
		\includegraphics[width=0.45\textwidth]{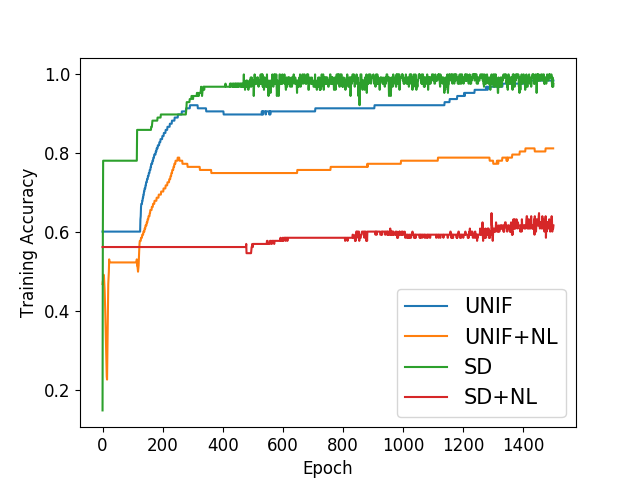} & 
		\includegraphics[width=0.45\textwidth]{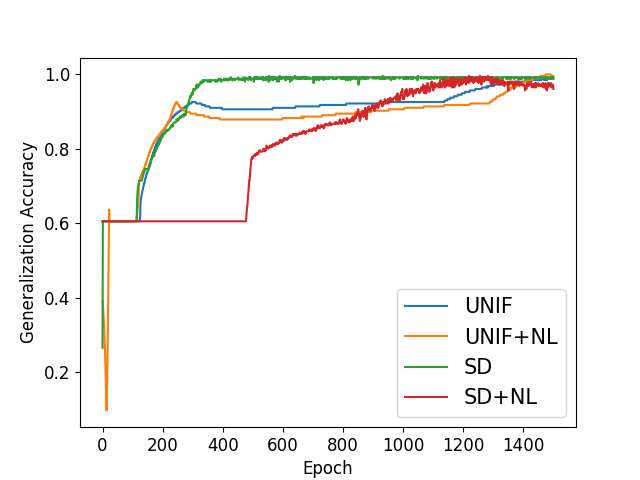}
	\end{tabular}
	\caption{The training and generalization accuracy versus epoch for UNIF, UNIF+NL, SD, SD+NL in the one-dimensional example, respectively. }\label{fig:1dacc}
\end{figure}

\begin{figure}[t]
	\subfigure[UNIF]
	{\begin{minipage}[t]{0.24\linewidth}
			\centering
			\includegraphics*[scale = .20]{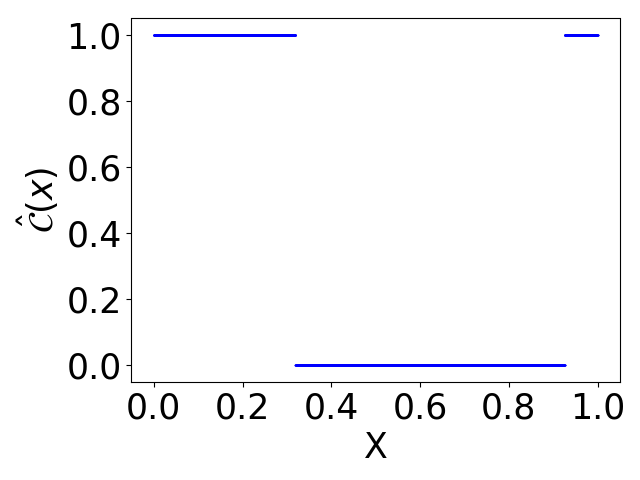}
	\end{minipage}}
	\subfigure[UNIF+NL]
	{\begin{minipage}[t]{0.24\linewidth}
			\centering
			\includegraphics*[scale = .2]{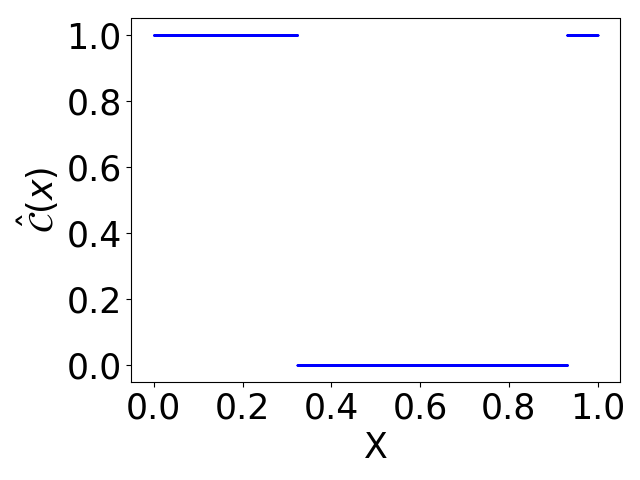}
		\end{minipage}}
	\subfigure[ SD]
	{\begin{minipage}[t]{0.24\linewidth}
			\centering
			\includegraphics*[scale = .2]{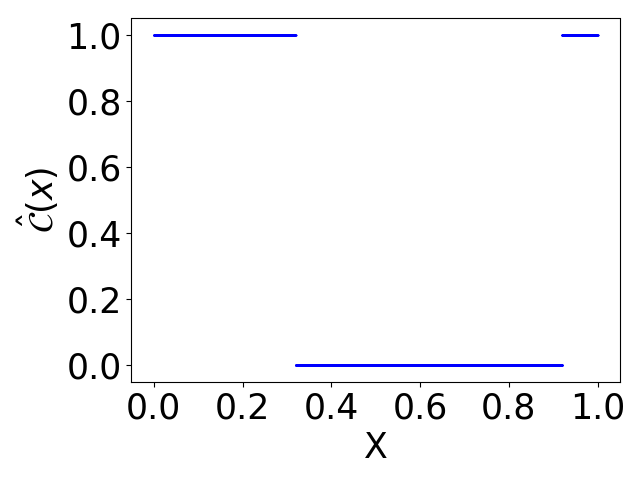}
		\end{minipage}}
	\subfigure[SD+NL]
	{\begin{minipage}[t]{0.24\linewidth}
			\centering
			\includegraphics*[scale = .2]{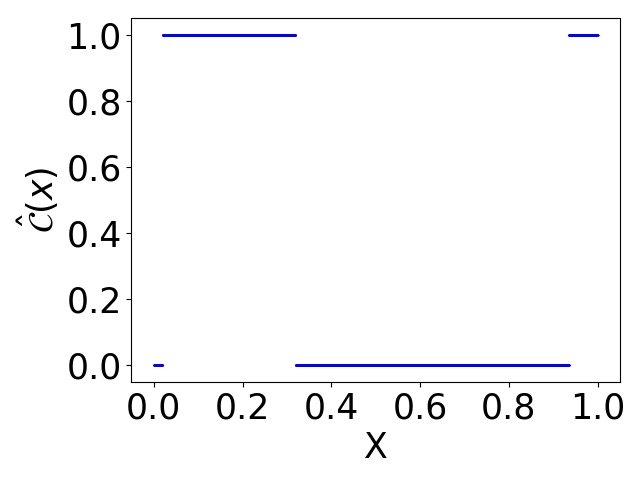}
		\end{minipage}}

	{\begin{minipage}[t]{0.24\linewidth}
			\centering
			\includegraphics*[scale = .2]{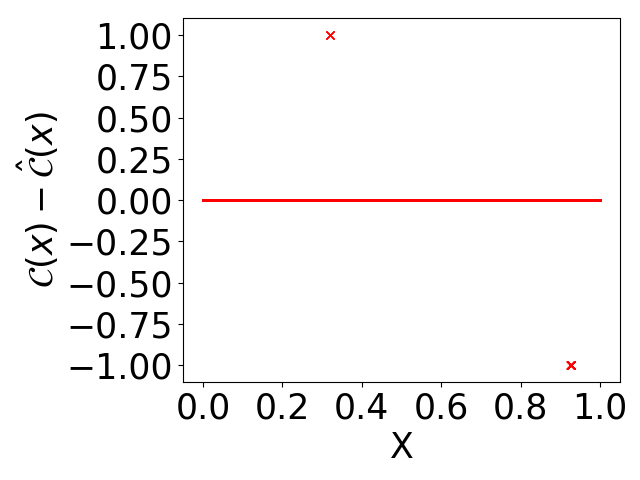}
	\end{minipage}}
	{\begin{minipage}[t]{0.24\linewidth}
			\centering
			\includegraphics*[scale = .2]{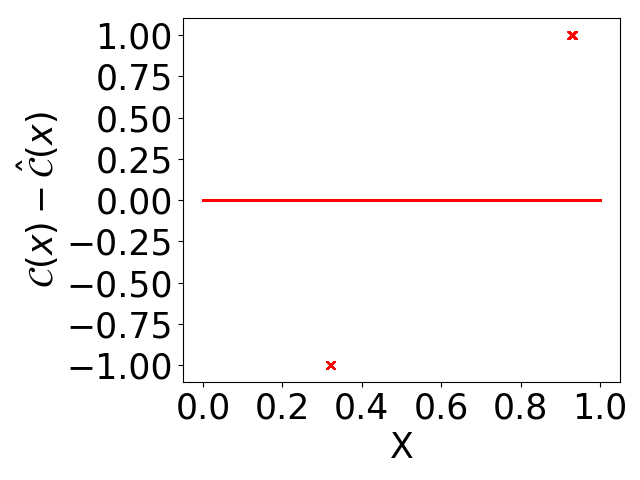}
		\end{minipage}}
	{\begin{minipage}[t]{0.24\linewidth}
			\centering
			\includegraphics*[scale = .2]{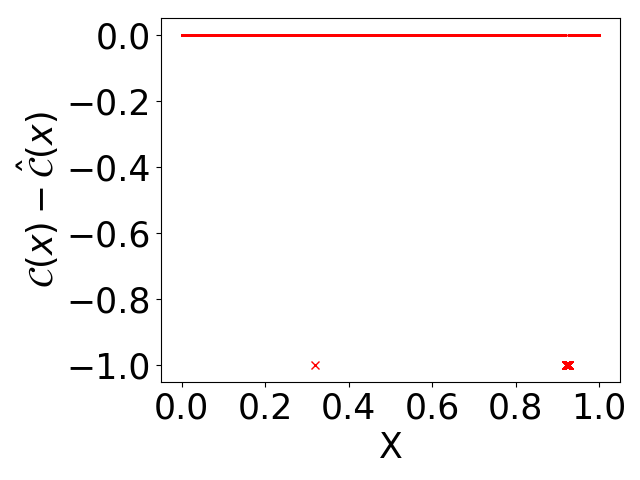}
		\end{minipage}}
	{\begin{minipage}[t]{0.24\linewidth}
			\centering
			\includegraphics*[scale = .2]{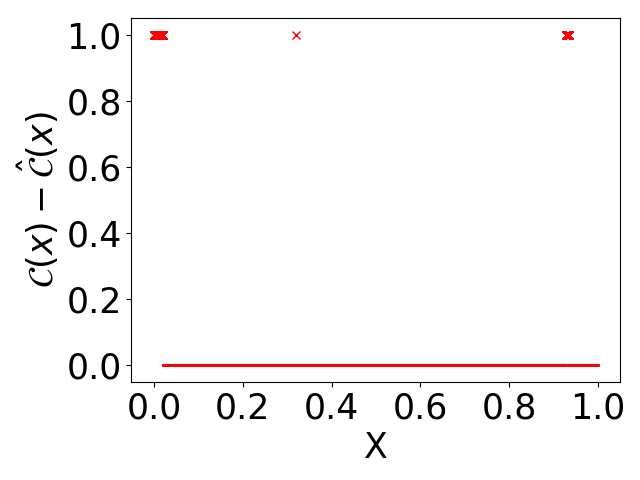}
		\end{minipage}}		
	
	\caption{Predicted classifiers $\hat \CC(x)$ (blue) and corresponding difference from true classifier $\CC(x) - \hat{\CC}(x)$ (red) by UNIF, UNIF+NL, SD, SD+NL using a size of $M=128$ training data for one-dimensional example. The wrongly predicted classifiers are marked as `x' in subfigures in the second row.}\label{fig:1d}
\end{figure}

Below, we also plot the training and generalization accuracy versus epoch in Figure~\ref{fig:1dacc}. The predicted ranking classifiers and corresponding difference from true values are given in Figures~\ref{fig:1d} for UNIF, UNIF+NL, SD, and SD+NL using a size of $M=128$ training data.

One can also notice that, in Figure~\ref{fig:1d}, the network predicts wrong classifiers not only near the boundaries but also at the ``fake'' boundary $x=0$ where the two response surfaces are close to each other. This is because, the training data points generated by SD are often near the boundaries or ``fake'' boundaries, and using noisy labels will lead to the points near boundaries or ``fake'' boundaries  having wrong classifiers, which makes the networks predict wrong classifiers at these places.

\subsection{Two-dimensional example}
In this subsection, we further study the sensitivity of deep learning algorithms to noisy labels, sampling locations and budget by a two-dimensional (2D) example used in \cite{HuLudkovski17}.  It treats a more complex setting with $L=5$ surfaces and a 2D input space $\XX=[-2,2]^2$, with a constant homoscedastic observation noise $\eps_\ell(x_1,x_2) \sim \mathcal{N}(0, \sigma_\ell^2)$,  $\sigma_\ell=0.5, \; \forall \ell=1,\cdots,5$.  The specific response functions for each surface and true classifier $\CC$ of problem \eqref{def_cal} is shown in Figure~\ref{fig:2d_fun}.

	\begin{minipage}{\textwidth}
		\begin{minipage}[t]{0.45\textwidth}
			\vspace{-125pt}
			\begin{equation}
			\begin{array}{lr}
				\hline
				\\[-0.8em]
				\text{Surface}& \text{Response}  \\ \hline\hline
				\\ [-0.8em]
				\mu_1(x_1,x_2) &  2-x_1^2 - 0.5 x_2^2 \\
				\\[-0.8em]
				\mu_2(x_1,x_2) &  2(x_1-1)^2 + 2x_2^2 -2 \\
				\\[-0.8em]
				\mu_3(x_1,x_2) &  2 \sin(2x_1)+2  \\
				\\[-0.8em]
				\mu_4(x_1,x_2) &  8(x_1-1)^2 + 8x_2^2 -3 \\
				\\[-0.8em]
				\mu_5(x_1,x_2) &  0.5(x_1+3)^2 +16x_2^2 -6  \\ \hline
				\end{array}
				\end{equation}
		\end{minipage}%
		\hspace{20pt}
			\begin{minipage}[t]{0.45\textwidth}
				\centering\includegraphics[width=0.9\textwidth]{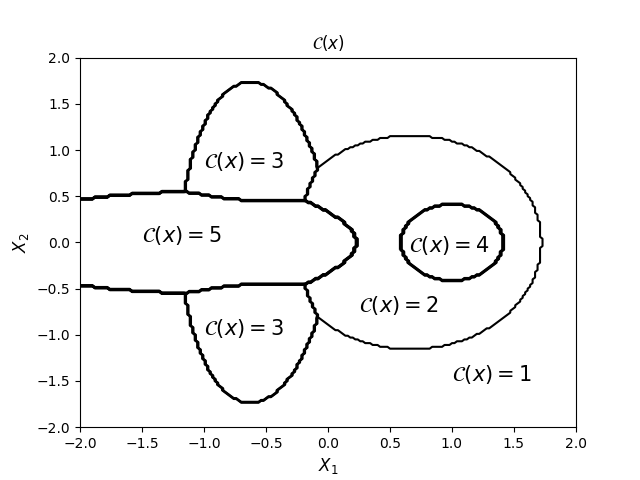}
			\end{minipage}
				\captionof{figure}{ Left: specific response functions for each surface; Right: the true ranking classifier for the two-dimensional example which divides the entire input space $[-2, 2]\times[-2, 2]$ into six parts.}\label{fig:2d_fun}		
	\end{minipage}

\begin{table}[H]
	\centering
	\caption{Training accuracy versus generalization accuracy for the 2-D example with different computational budget $M$. The acronyms used are: UNIF = uniform grids on $\mathcal{X}$, SD = grids generated by Gap-SUR in \cite{HuLudkovski17}, NL = training with noisy labels. \label{tbl:2d}}
	\begin{tabu}{|l|cc|cc|cc|}
		\hline
		{Method/Budget} & \multicolumn{2}{c|}{\textbf{M = 256}}   & \multicolumn{2}{c|}{\textbf{M = 576}}  & \multicolumn{2}{c|}{\textbf{M = 1024}}   \\ 
		& Train. Acc. & Gen. Acc. & Train. Acc. & Gen. Acc. & Train. Acc. & Gen. Acc. \\ \tabucline[1pt]{1-7}
		UNIF &99.9\% & 94.7\%&99.7\%&96.6\%&99.5\%& 97.7\% \\
		UNIF + NL &98.4\%&92.8\%&93.2\%&95.1\%&90.8\%&96\%\\ \hline
		SD&96.9\%&94.4\%&96.1\%&96.4\%&96.1\%& 97.4\% \\
		SD + NL &82.0\%&94.1\%&71.8\%&94.6\%&66.8\%&96.8\%\\ \hline	
	\end{tabu}
\end{table}

\begin{figure}[t]
	\begin{tabular}{cc}
		\includegraphics[width=0.45\textwidth]{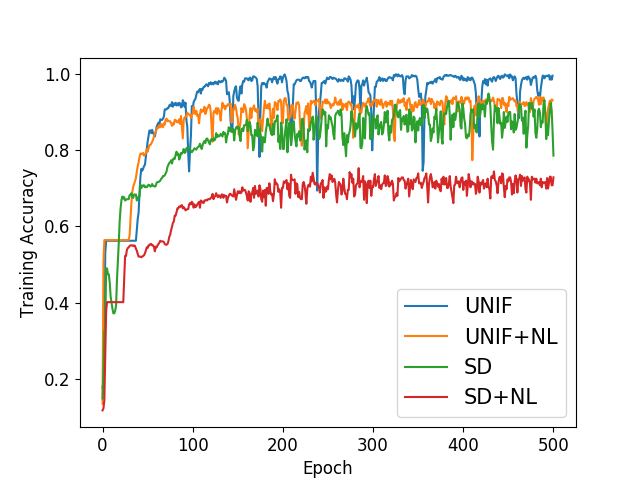} & 
		\includegraphics[width=0.45\textwidth]{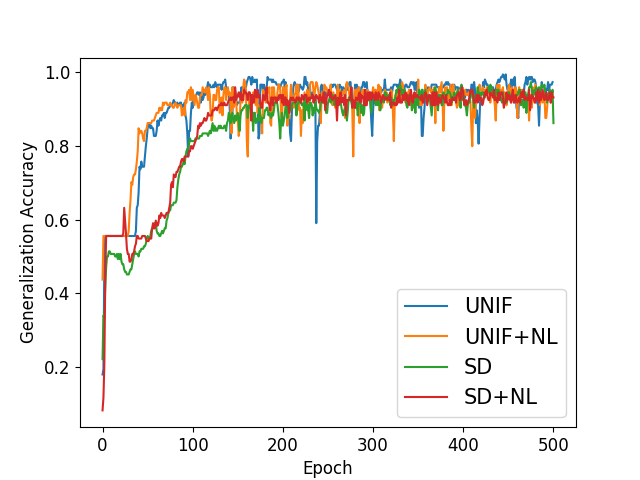}
	\end{tabular}
	\caption{The training and generalization accuracy versus epoch for UNIF, UNIF+NL, SD, SD+NL in the two-dimensional example, respectively. }\label{fig:2dacc}
\end{figure}

\begin{figure}[t]
	\subfigure[Predicted classifier by UNIF]
	{\begin{minipage}[t]{0.5\linewidth}
			\centering
			\includegraphics*[scale = .4]{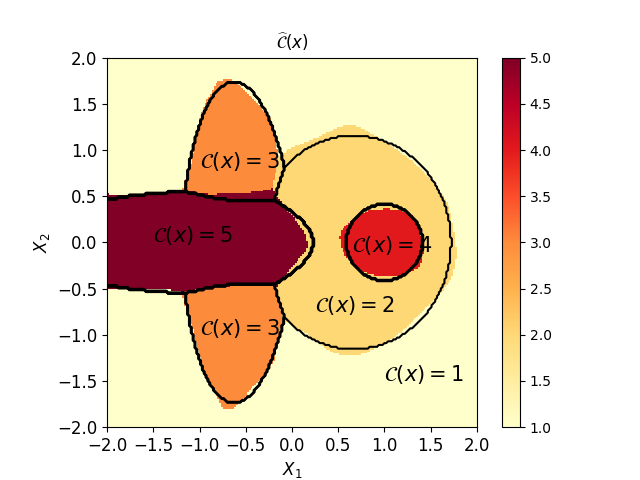}
	\end{minipage}}
	\subfigure[Predicted classifier by UNIF+NL]
	{\begin{minipage}[t]{0.5\linewidth}
			\centering
			\includegraphics*[scale = .4]{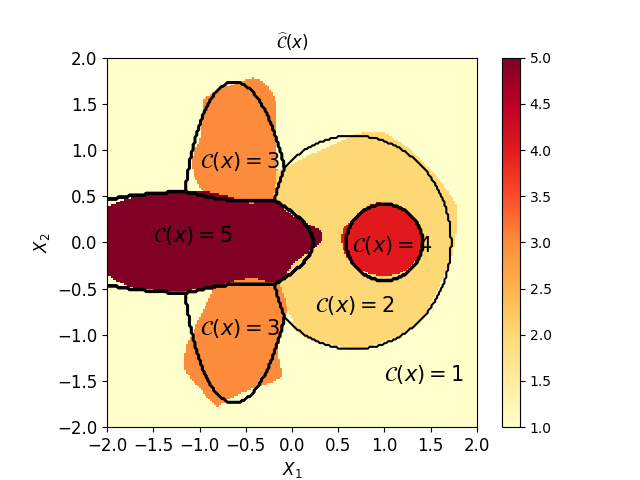}
	\end{minipage}}
		\subfigure[Predicted classifier by SD]
		{\begin{minipage}[t]{0.5\linewidth}
				\centering
				\includegraphics*[scale = .4]{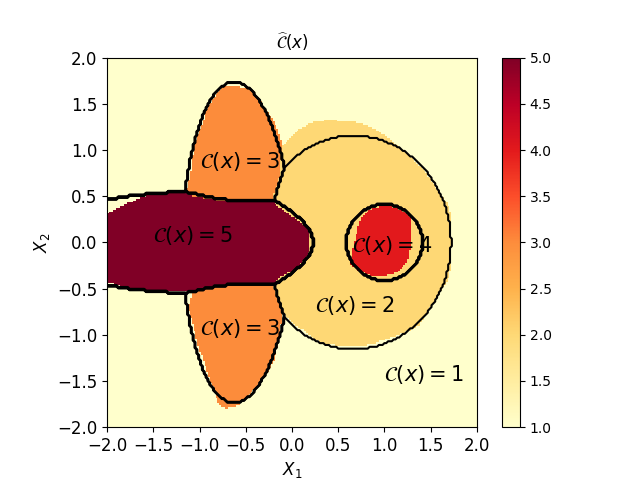}
			\end{minipage}}
			\subfigure[Predicted classifier by SD+NL]
			{\begin{minipage}[t]{0.5\linewidth}
					\centering
					\includegraphics*[scale = .45]{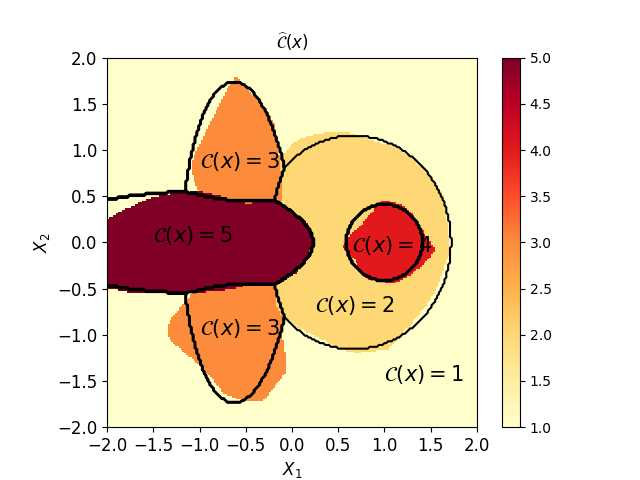}
				\end{minipage}}
	\caption{Predicted 2-D ranking on $\XX = [-2,2]^2$ using different designs: UNIF (top-left), UNIF+NL (top-right), SD (bottom-left) and SD+NL (bottom-right). The solid black lines show the true $\CC(x_1, x_2)$, the colored regions show the estimated classifiers $\hat{\CC}_i$ for M = 576. }\label{fig:2d}
\end{figure}

We perform the same four types of training data (cf. Table~\ref{tbl:datadesign}) as in the 1D example, and show the results of accuracy in Table~\ref{tbl:2d}. We also plot the training and generalization accuracy versus epoch in Figure~\ref{fig:2dacc}. The predicted ranking classifiers and corresponding difference from true values are given in Figures~\ref{fig:2d} for UNIF, UNIF+NL, SD, and SD+NL using a size of $M=576$ training data. These testing results show a consistent conclusion with the 1D example, {\it i.e.},  the generalization accuracy is not very sensitive to the noise in the training data set; using SD for generating sampling locations $x^{1:M}$ potentially decrease the training accuracy due to more errors in the learning data set, and increasing the budget $M$ will make it even lower before points around the boundaries get saturated.

\begin{figure}[t]
	\centering\includegraphics[width=0.6\textwidth]{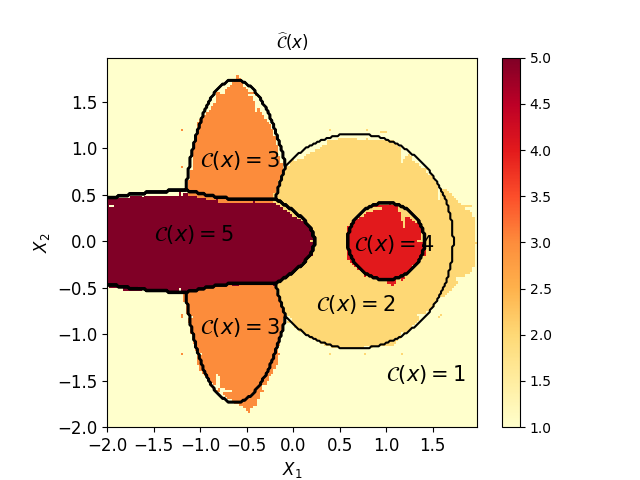}
	\caption{Predicted 2-D ranking classifiers on $\XX = [-2, 2]^2$ using UNet. The solid black lines are the true classifier $\CC(x_1, x_2)$, the colored regions indicates the estimated minimal index using $M = 576$ training data.}\label{fig:UNet}
\end{figure}

To our best knowledge, there are no theoretical results in literature indicating which type of architecture works best for specific applications. Instead, one usually implements a few architectures of NNs, and choose the best of them empirically. Remark that, considering ranking response surfaces as image segmentation allows one to use a broad class of deep neural networks. In this example, we implement the UNet architecture (right panel in Figure~\ref{fig:arch}) using $M=576$ uniform grid points with noisy labels, and show the predicted classifier in Figure~\ref{fig:UNet}. The UNet has a generalization accuracy of $96.44\%$, presenting a better performance than feed-forward NNs (corresponding to the $95.1\%$ in Table~\ref{tbl:2d}), with comparable training time. It is also visualized by comparing Figure~\ref{fig:UNet} to Figure~\ref{fig:2d} (b) (d).


\subsection{Ten-dimensional example}

To show the robustness of deep learning algorithms, we consider a ten-dimensional synthetic example in this section, for which using the method in our previous work \cite{HuLudkovski17} will have a problem of unaffordable computational time. For an easiness of comparison, we construct the example as follows, which the true classifier $\MCC$ is accessible.

Let $L = 3$ surfaces and $\mc{X} = [-1,1]^d$ with $d = 10$. The surfaces we use include embedded Hartmann 6-D function, rescaled Styblinski-Tang function, and rescaled Trid function as described in Table~\ref{tbl:10d_func}. The Hartmann 6-D function has 6 local minima and a global minima at $$\bm x^\ast = (0.20169, 0.150011, 0.476874,0.275332,0.311652,0.6573).$$ We embed this function to ten-dimensional space as our $\mu_1(\bm x)$. The original Styblinski-Tang function $f(\bm x) = \half \sum_{i=1}^d x_i^4-16x_i^2 + 5x_i$ is usually evaluated on $[-5,5]^d$ with a global minimum at $\bm{x}^\ast = (-2.903534, \ldots, -2.903534)$. We rescale the domain and the function to make it comparable with Hartmann 6. The same applies to the Trid function. We study the example using different layers of NNs trained by both clean and noisy data, and present the training and generalization accuracy in Table~\ref{tbl:10d}, where the first number is the training accuracy, and the second in parenthesis is the generalization accuracy. We notice that, similar to previous 1-D and 2-D examples, when the NNs are trained by clean data, the training accuracy is better than the generalization accuracy as in standard deep learning theory, while when the NNs are trained by noisy data, the generalization accuracy is better. This is due to a fact that, when there are errors (caused by noise) in the training data set, the NNs auto-detect these errors and avoid learning from them, which decreases the training ``accuracy'', while the used generalization data set only contains clean data.

\begin{table}[t]
	\centering
	\caption{The three response surface functions in the ten-dimensional example.}\label{tbl:10d_func}
\begin{equation*}
\begin{array}{ll}
\hline
\\[-0.8em]
\text{Surface}& \text{Response}  \\ \hline\hline
\\ [-0.8em]
\text{Hartmann 6-D }& \mu_1(\bm x) = -\sum_{i=1}^4 \alpha_i \exp\left(-\sum_{j=1}^6 A_{ij}(x_j-P_{ij})^2\right),
\\ \vspace{5pt}
& \text{with }\alpha = \begin{bmatrix}
1\\1.2\\3.0\\3.2
\end{bmatrix},  
A = \begin{bmatrix}
10     & 3 & 17 & 3.50& 1.7 & 8 \\
0.05     & 10&17& 0.1 & 8 & 14 \\
3&3.5&1.7&10&17&8 \\
17&8      & 0.05& 10 &0.1& 14
\end{bmatrix},
\\ 
&\qquad P = \begin{bmatrix}
0.1312    & 0.1696 & 0.5569 & 0.0124& 0.8283 & 0.5886 \\
0.2329     & 0.4135&0.8307& 0.3736 & 0.1004 & 0.9991 \\
0.2348&0.1451&0.3522&0.2883&0.3047&0.6650 \\
0.4047&0.8828     & 0.8732& 0.5743 &0.1091& 0.0381
\end{bmatrix};
\\

\\[-0.8em]
\text{Styblinski-Tang } &  \mu_2(\bm x)= \frac{1}{2d}\sum_{i=1}^d 625x_i^4-400x_i^2 + 25x_i ; \\ 
\\[-0.8em]
\text{Trid } &  \mu_3(\bm x) = \half\left(\sum_{i=1}^d (x_i-1)^d - \sum_{i=2}^d x_ix_{i-1}\right)-5 . \\
\hline
\end{array}
\end{equation*}
\end{table}

\begin{table}[t]
	\centering
	\caption{Training accuracy versus generalization accuracy for the 10-D example using different layers of neural networks (NNs) trained by both clean and noisy data. The first number is the training accuracy, and the second in parenthesis is the generalization accuracy. Note that, when the NNs are trained by clean data, the training accuracy is better than generalization accuracy as expected, while when the NNs are trained by noisy data, the generalization accuracy is better. This is due to a fact that, when there are errors (caused by noise) in the training data set, the NNs auto-detect these errors and avoid learning from them, which decreases the training ``accuracy'', while the used generalization data set only contains clean data.}\label{tbl:10d}
	\begin{tabular}{|c|c|c|c|}
		\hline
		{Noise Level/\# of Layers} & 2 Layers   & 3 Layers  & 4 Layers\\  \hline
		No noise & 92.5\% (92.1\%)  & 94.8\% (93.9\%) &94.5\% (92.7\%) \\ \hline
		$\sigma = [0.5,0.4,0.45]$&90.5\% (91.1\%)&91.8\% (91.8\%)&93.1\%(92.4\%) \\ \hline
		$\sigma = [0.7,0.75,0.8]$ &89.6\% (91.4\%)&91.0\% (92.9\%)&91.6\% (92.1\%)\\ \hline
	\end{tabular}
\end{table}

\section{Bermudan option pricing}\label{sec:Bermudan}

An important problem in computational finance is pricing Bermudan/American-type options. It has been studied extensively in the literature, for instance,  via regression methods \cite{Carriere96, Longstaff, tsitsiklis2001regression,kohler2010review,Belomestny11,letourneau2014refining,GoMoZa:193,GoMoZa:194} and variance reduction \cite{juneja2009variance,hepperger2013pricing,jain2015stochastic}, primal-dual formulation \cite{andersen2004primal, broadie2008improved, Bender11}, adaptive experiment designs for optimal stopping \cite{GL13,ludkovski2015kriging} and counter-finding \cite{lyu2018evaluating}, to list a few.

For Bermudan-type options, the buyer has the right to exercise at a set number of times. Assume they are discretely spaced, denoted by  $\{t_0 = 0, t_1, t_2, \ldots, t_{N-1}, t_N = T\}$ bounded by the maturity date $T$, the price is determined by the maximum expected payoff over all possible $t_i$. To maximize (or optimize) buyer's profit, one wants to execute/stop the option in a way that the largest payoff will be received,  based on the information known up to today. Therefore, this can be formulated as an optimal stopping problem, and is typically solved by backward induction. In other words, one first computes a value function at the expiration date, and then recursively works backward, computing value functions and making a decision on whether to execute for preceding periods. At each possible executing time $t_i$, the decision space is small, containing only two elements $\mk{L} = $\{stop, continue\}, which makes it a natural application of ranking problems. Therefore, in this section, we apply the deep learning algorithms to price Bermudan-type options.

Let $X_t \in \XX \subset \RR^d$ be the price of underlying assets,  $\FF_n = \sigma(X_{t_{0:n}})$ be the $\sigma-$algebra generated by $(X_{t_i})_{i=1}^n$ and $\mc{S}$ be the collection of all $(\FF_n)$ stopping times. Then pricing Bermudan option is essentially to maximize the expected reward $h(\tau, X_\tau)$ over all stopping times $\tau \in \mc{S}$. Mathematically, denoting by $V(t,x)$ the value of such an option at time $t$ with current price $X_t = x$:
\begin{equation}
V(t,x) := \sup_{\tau \geq t, \tau \in \mc{S}} \EE_{t,x}[h(\tau, X_\tau)],
\end{equation}
the price is given by $V(0,X_0)$. By dynamic programming, 
\begin{equation}
V(t_i,x) = \max\{h(t_i, x), C_V(t_i, x)\},
\end{equation}
where $C_V(t_i, x)$ is the continuation value, corresponding to the action ``{continue}'' in $\mk{L}$:
\begin{equation}\label{def_cont}
C_V(t_i, x) :=  \EE_{t_i, x}[V(t_{i+1}, X_{t_{i+1}})],
\end{equation}
and $h(t_i,x)$ is the immediate payoff if one chooses to exercise the option at $t_i$, related to the action ``{stop}'' in $\mk{L}$. Denote by  $\tau^\ast(t,x)$ the stopping time when the supremum is attained, it is identified by 
\begin{equation}
\{\tau^\ast(t_i,x) = t_i\} = \{x \in \XX: h(t_i, x) \geq C_V(t_i,x)\}.
\end{equation}
Using the above formulation, one can estimate $\tau^\ast$ recursively from $t_N$ to $t_0$. 

Rephrasing it in terms of our ranking problem setup, fixing time $t_i$, the choice between ``stop'' or ``continue'' is equivalent to find $\CC(t_i, x) := \arg\max \{\mu_{stop}(t_i, x), \mu_{cont}(t_i, x)\}$ over $\XX$, a segmentation of the input space between continuation and stopping regions. Here $\mu_{stop} = h$ can be evaluated deterministically, while a closed-form formula for $ \mu_{cont} = C_V$ is typically not available, but accessible through simulations. Moreover, its evaluation also depends on all further classifiers $\CC(t_j, x)$, $i+1 \leq j \leq N$. To be more precise, for a given collection of estimated classifier $\hat{\CC}(t_{i+1:N}, \cdot)$, define the pathwise stopping strategy:
 \begin{equation}\label{def_tau}
 \hat{\tau}(t_i,x)(\omega) := \inf\{t_j > t_i: \hat\CC(t_j, X_{t_j}(\omega)) = stop\} \wedge T,
 \end{equation}
 for every path $X_{(\cdot)}(\omega)$ with initial position $X_{t_i} = x$. Now, by simulating $R$ independent paths $x^r_{t_{i:N}}$ starting from $x^r_{t_i} = x$, $r = 1, \ldots, R$, the continuation value is estimated by
 \begin{equation}\label{def_approxcont}
\hat C_V(t_i, x) := \frac{1}{R} \sum_{r=1}^R h(\hat\tau(t_i, x_{(\cdot)}^r), x^r_{\hat{\tau}(t_i, x_{(\cdot)}^r)}),
 \end{equation}
 producing the estimated classifier $\hat \CC(t_i, \cdot)$ at location $x$
 \begin{align}\label{def_cal_fm}
 \hat\CC(t_i, x) &:= \arg\max\{\mu_{stop}(t_i, x), Y_{cont}(t_i, x)\} , \\
 Y_{cont}(t_i, x) &:= C_V(t_i, x) + \eps_{cont}(t_i, x), \quad \eps_{cont} := \hat C_V - C_V. 
 \end{align}
The term $\eps_{cont} = \hat C_V - C_V$ summarizes the simulation noise from two approximations: the usage of estimated classifiers $\hat \CC(t_{i+1:N}, \cdot)$ that determines the exercise strategy \eqref{def_tau} and  the action of replacing the expectation in \eqref{def_cont} by the average over sample paths \eqref{def_approxcont}. Once the decision maps $\hat \CC(t_{1:N}, \cdot)$ are obtained, $V(0,X_0)$ is estimated on a out-of-sample set of $M'$ realizations of $X_{(\cdot)}$.

We use deep learning algorithms for the purpose of efficient and accurate learning of $\hat \CC$ over the entire input $\XX$ through finite samples. Starting from the maturity time $t_N = T$, the learning is done recursively back to $t_0=0$, with each $\hat{\CC}(t_i, \cdot)$ produced by a neural network. In practice, one can improve the label quality in the training set by increasing the number of simulations $R$ which can reduce the magnitude of $\eps_{cont}$. However, we remark that errors are tolerated and $R$ does not need to be too large, as long as they do not affect the ordering of $Y_{cont}$ and $h$. We describe the pricing procedure in Algorithm~\ref{def_algorithm}.

\begin{algorithm}[t]
	\caption{Deep Learning for Pricing Bermudan Option \label{def_algorithm}}
	\begin{algorithmic}[1]
		\REQUIRE $M$ = \# of sampling locations, $R$ = \# of sample paths at each grid, $M'$ = \# of out-of-sample paths for pricing, $X_0$ = initial price 
		\STATE Define the classifier at maturity $t_N = T$:  $\hat \CC(t_N, x) = stop$
		\FOR{$i \gets N-1$ downto $1$}
		\STATE Generate sampling locations $x \in \XX$ of size $M$
		\FOR{each location $x$}
		\STATE Sample $R$ paths $x^r_{t_{i:N}}$ with $x^r_{t_i} = x$, $r = 1, 
		\ldots, R$
		\STATE Construct the pathwise stopping strategy  $\hat{\tau}^r(t_i, x) := \inf\{t_j > t_i: \hat\CC(t_j, x^r_{t_j}) = stop\}$ \label{algo_strategy}
		\STATE Compute the continuation value by $\hat C_V(t_i, x) := \frac{1}{R} \sum_{r=1}^R h(\hat\tau^r(t_i, x), x^r_{\hat{\tau}^r(t_i,x)})$ \label{algo_cont}
		\IF{$ \hat C_V(t_i, x) > h(t_i, x)$}
		\STATE $z(x) \leftarrow continue$
		\ELSE
		\STATE $z(x) \leftarrow stop$
		\ENDIF
		\ENDFOR
		
		\STATE Train a neural network with the previously generated samples of $(x, z(x))$ as the input and the classifier $\hat \CC(t_i, \cdot)$ as the output 
		\ENDFOR
		
		\STATE Generate $M'$ out-of-sample paths $x^r_{t_{0:N}}$ with $x_{t_0}^r = X_0$, $r = 1, \ldots, M'$
		\STATE Compute $\hat C_V(t_0, X_0)$ by repeating Step \ref{algo_strategy} and \ref{algo_cont} for $(t_0, X_0)$
		\RETURN Estimated price  $\hat V(t_0, X_0) = \max\{h(t_0, X_0), \hat C_V(t_0, X_0)\}$.
	\end{algorithmic}
\end{algorithm}

Let us also remark that, this specific application of ranking response surfaces by deep learning actually has the same setup as the problem studied in \cite{BeChJe:18}, where the backward recursive stopping decisions are approximated by a feed-forward NN. In some sense, \cite{BeChJe:18} uses the feed-forward NN as interpolation to approximate the function with values $0$ and $1$ representing ``continuation'' and ``stopping''. In our work, by recasting the problem as image segmentation, one is allowed to use the more delicate architecture of neural networks ({\it e.g.} UNet), which increases computational efficiency.


With all above efforts on reformulating Bermudan option pricing as ranking response surfaces, we perform a numerical study of 2-D max-Call $h(t,x) = e^{-rt}(\max(x_1, x_2) - K)_+$.  The underlying assets $X = (X_1, X_2)$ are modeled by geometric Brownian motions,
\begin{equation}\label{def_Xi}
\ud X_i(t) = (r-\delta) X_i(t) \ud t + \sigma X_i(t) \ud W_i(t), \quad i = 1, 2,
\end{equation}
where $(W_1, W_2)$ is a 2-D standard BM, with the parameters from \cite{andersen2004primal}:
\begin{equation}\label{def_parameter}
r = 5\%, \quad \delta = 10\%, \quad \sigma = 20\%, \quad X(0) = (90, 90), \quad K = 100, \quad T = 3, \quad N = 9, \quad t_i = i\frac{T}{N} .
\end{equation}

\begin{figure}[t]
	\begin{tabular}{cccc}
		\includegraphics[width=0.23\textwidth]{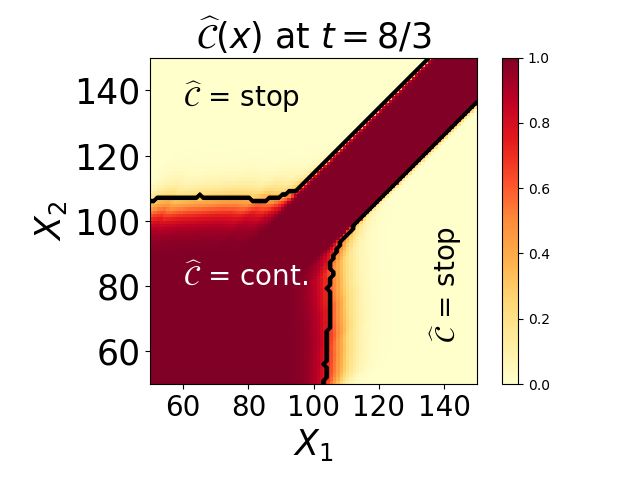} & 
		\includegraphics[width=0.23\textwidth]{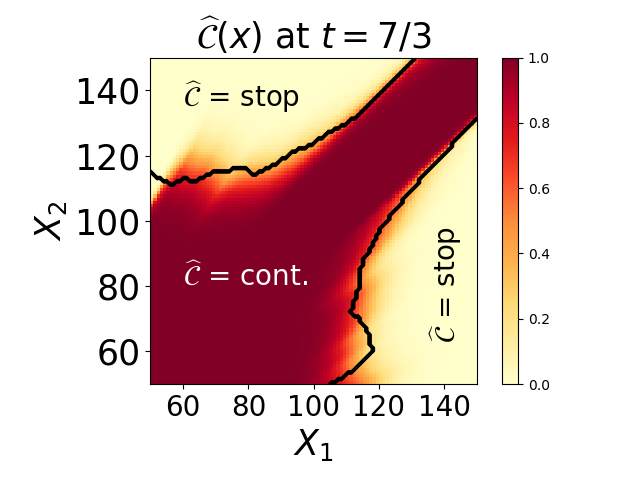} &
		\includegraphics[width=0.23\textwidth]{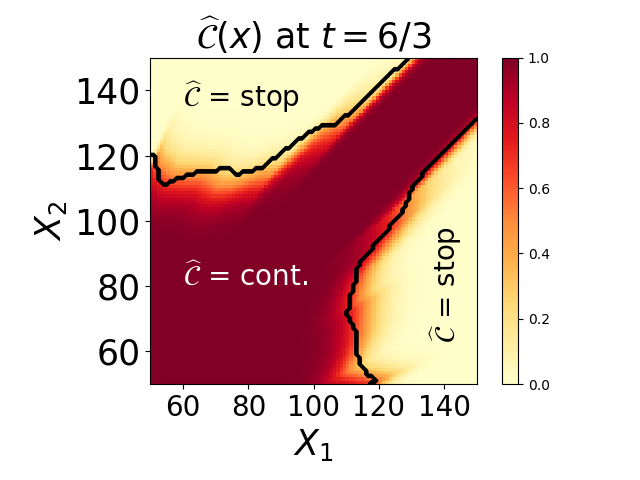} & 
		\includegraphics[width=0.23\textwidth]{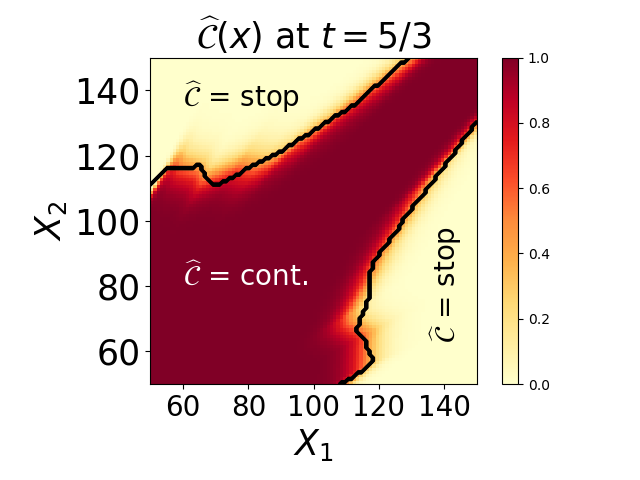} \\
		\includegraphics[width=0.23\textwidth]{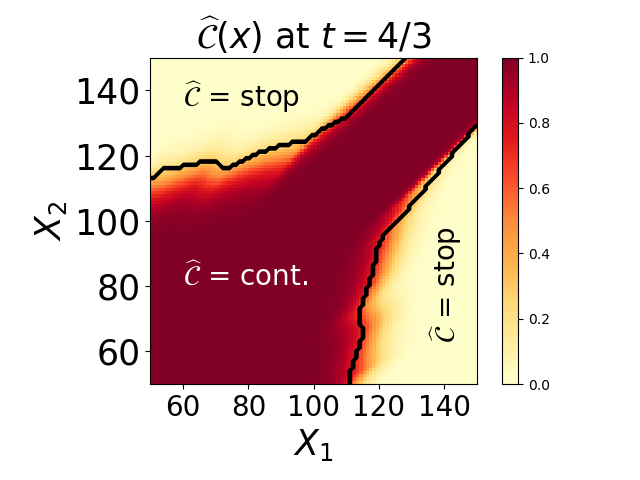} & 
		\includegraphics[width=0.23\textwidth]{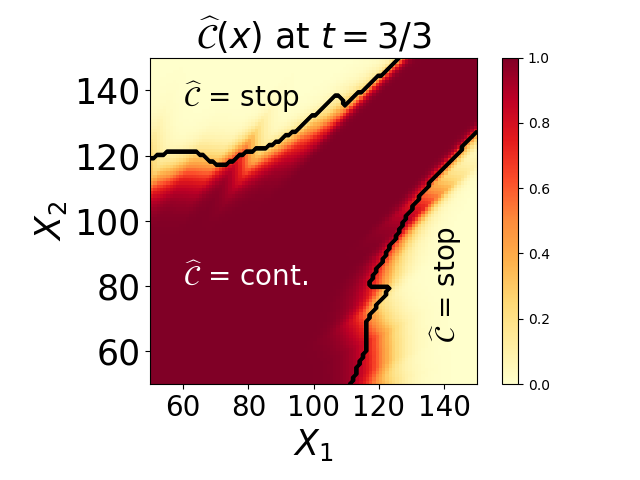} &
		\includegraphics[width=0.23\textwidth]{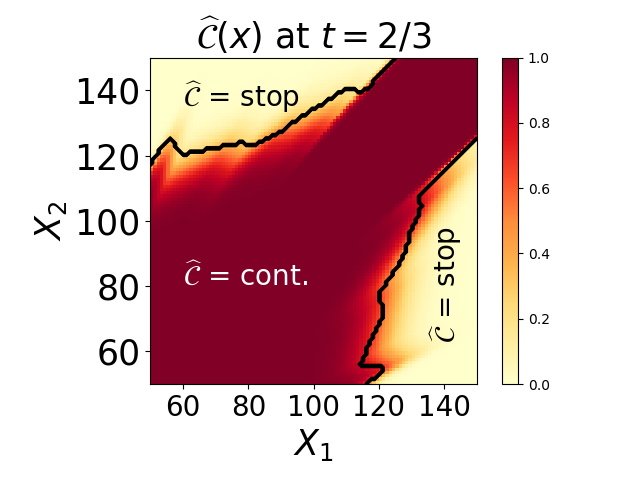} & 
		\includegraphics[width=0.23\textwidth]{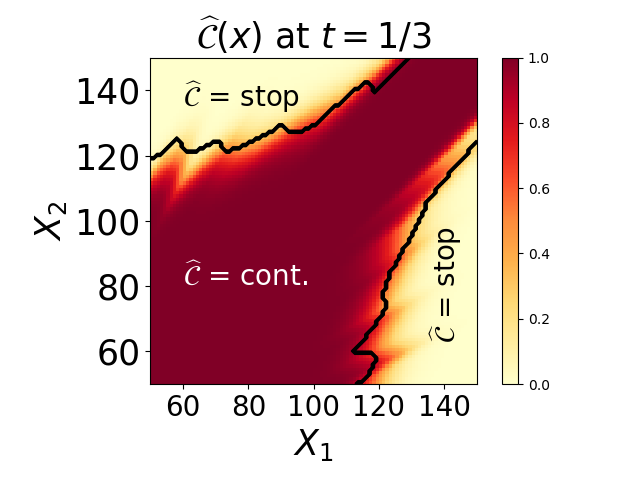}
	\end{tabular}
	\caption{Deep learning of the decision maps $\hat \CC(t_i, \cdot)$ with the training data generated by UNIF+NL (cf. Table~\ref{tbl:datadesign}). Black solid lines show the estimate boundaries of $\{\text{continue}, \text{stop}\}$.  The colorbar indicates the probability that the  neural network outputs for the decision of ``continuation''. Darker color means more confidence of the neural network feels about the classification.}\label{fig:fm}
\end{figure}

From the numerical tests in Section~\ref{sec:numerics}, one can notice that the generalization accuracy of UNIF (or UNIF+NL) is higher than that of SD (or SD+NL). Moreover, implementation of the UNet on uniform grids is easier than using the points generated by sequential design. Therefore, we will only use neural networks trained by data generated on uniform grids for computing the Bermudan option price. Figure~\ref{fig:fm} shows the decision maps estimated recursively by deep learning algorithms for different time slices. They are plotted backward in time, in the order of being generated. The trivial decision map at $t_N = 3$ is not included. A map at time $0$ is not needed, as we know $X(0)$. $\hat{\CC}$ are generated by a uniform $32\times 32$ grids on $[50, 150]^2$ with $R = 100$ replication at each location. In the red area, the continuation value is higher, and thus it is optimal to keep holding the option; while in yellow regions, immediate rewards are higher, making it optimal to exercise right away. The estimated price $\hat V(0, X(0))$ is 8.05 with a standard deviation 0.029, based on $M' = 160,000$ out-of-sample paths repeating 100 times. This estimation is quite close to the true value of the option $8.075$, computed using a 2-D binomial lattice for this case \cite{Belomestny11}. 

{Next, we  investigate the performance of deep learning algorithms on pricing $d$-dimensional max-Call Bermudan option, that is, the payoff function is $h(t,\bm x) = e^{-rt}(\max(x_1, \ldots, x_d)-K)_+$, $\bm x \in \RR^d$. The underlying assets $X_i(t)$ follow dynamics \eqref{def_Xi}, for $i = 1, \ldots, d$,  and the parameters follow the choice \eqref{def_parameter}. The decision maps are trained by $1024\times d$ samples on $[30, 180]^d$ with $R = 100$ replication at each location. The samples are generated using Latin Hypercube, ensuring that the set is representative. In Table~\ref{tbl:fm}, we list the computed estimated price $\hat V(0, X(0))$ and simulation time for various choices of parameters and dimensions, where our results agree well with the binomial lattice method \cite{boyle1989numerical} and primal-dual method \cite{andersen2004primal, broadie2008improved}. One can also observe that, the proposed algorithms based on deep learning can easily simulate examples of high dimensions (the highest $d=50$ in our simulations), while conventional methods usually become infeasible for examples of dimension$>5$.}

\begin{table}[t]

	\centering
	\caption{Results for max-call options on $d$ underlying assets with parameter chosen in \eqref{def_parameter}. The computational time are in seconds. The binomial value, and BC 95\% CI are obtained from \cite{andersen2004primal} and \cite{broadie2008improved} respectively, which are only available for low dimensional cases ($d\leq 5$). }\label{tbl:fm}
	\begin{tabular}{|c|c|c|c|c|c|c|}
		\hline
		d & $X(0)$   & $\hat V(0, X(0))$  & Training Time & Generalization Time & Binomial & BC 95\% CI \\  \hline
		2 & 90 & 8.065 & &74.412 & 8.075 & \\
		2 & 100 & 13.897 & 303.03&75.714 & 13.902&\\
		2 & 110 & 21.349 & &74.195 & 21.345&\\ \hline
		
		3 & 90 & 11.271&& 76.366 & 11.29& \\
		3 & 100 & 18.659&768.33& 78.596& 18.69 & \\
		3 & 110 & 27.606 && 75.895& 27.58 & \\ \hline 
		
		5 & 90 & 16.594&& 78.169 & & [16.620, 16.653]\\
		5 & 100 & 26.126 &912.20&75.143 &  &[26.115, 26.164] \\
		5 & 110 & 36.734&& 76.267 &  & [36.710, 36.798]\\ \hline 
		
		10 & 90 & 26.220&& 80.497& & \\
		10 & 100 & 37.329 &1025.19& 79.402&  & \\
		10 & 110 & 50.557&& 79.760 &  & \\ \hline 
		
		20 & 90 & 37.695 && 83.677& & \\
		20 & 100 & 51.676 &1695.68&84.529&  & \\
		20 & 110 & 65.734&& 84.978&  & \\ \hline 
		
		50 & 90 & 53.653&& 91.561 & & \\
		50 & 100 & 69.130 &2594.67&90.734&  & \\
		50 & 110 & 84.444&& 90.878&  & \\ \hline 
 	\end{tabular}
\end{table}

In Table~\ref{tbl:fm}, the fourth column lists the time of training the classifiers $\hat \CC(t_i, \cdot)$, and the fifth column gives the computational time of estimating price $\hat V(0, X(0))$ using out-of-sample paths. We remark that, firstly, the training only needs to be done once to price derivatives with different $X(0)$, and this is why we only report the training time for different $X(0)$ with the same dimension $d$.  Secondly, the training time is related to the random initialization of network parameters. The actual training time is affected greatly by the empirical stopping criteria measured by the cross-entropy loss function and how close the initialized network parameters are to the local minimizers. Thus, this part is only for reference, and one may significantly shorten it if one starts from the pre-trained parameters. This is related to the third remark: the advantage of treating the whole problem as a classification. By doing so, one can generalize to other problem settings by using the fine-tuning, which is common in deep learning. In practice, practitioners can pre-train the neural network to identify the stopping region on a large dataset ({\it i.e.} underlying assets with all different returns, volatilities, strike price and maturity), which may take a long time. But once this is done, and if the new dataset is not drastically different from the original dataset (which is very likely), the pre-trained network will already learn some features that are relevant to the current classification problem. Thus, only fine-tuning is needed, and here are a couple of common fine-tuning techniques: 1) truncating the last layer of the pre-trained network and replacing it with a new layer, after which people only train the new layer while freezing others';  and 2) using a smaller learning rate of the stochastic gradient descent algorithm, which is usually ten times smaller than the one used for scratch training. Therefore, the generalization time is more relevant in our algorithm, which is the major cost for derivative pricing after the initial pre-training investment.

\section{Conclusion and future works}\label{sec:conclusion}

By recasting ranking response surfaces as image segmentation, we propose to use deep neural networks (NNs) with various architectures ({\it e.g.}, feed-forward NNs, UNet) as computational tools. Specifically, we consider labeling the entire input space using the index of the minimal surface, which segments the input space into distinct parts and allows one to use deep neural networks for efficient computation. This gives an alternative way of efficiently solving the problem instead of using sequential design in our previous work \cite{HuLudkovski17}. In particular, deep learning algorithms provide a scalable model and make the predicted results no more depend on the assumptions used in Gaussian process (GP) metamodels, for example, the assumptions on local/non-local kernels. Moreover, considering ranking response surfaces as image segmentation allows one to use a broad class of neural networks({\it e.g.}, CNNs, UNet, and SegNet). Although based on what we know so far, there is no existing result in literature rigorously discussing which architecture works the best for image segmentation,  an allowance of a broader class of NNs will make one choose an architecture with better performance empirically. For example, in the two-dimensional synthetic test, UNet produces a better accuracy than feed-forward NNs with comparable training time. A few more examples including a ten-dimensional synthetic test and the Bermudan option pricing are presented to show the success of deep neural networks in ranking response surfaces, which makes it possible to tackle more complicated problems, {\it e.g.}, optimal stopping game. Noticing that samples around the partition boundaries usually has low signal-to-noise level and potentially increase the chance of mislabeling, a natural extension is to consider replication/batching at those locations. Recent work  \cite{binois2018replication} by Binois {\it et al.} addresses this issue using GP metamodels, and we plan to study this problem by deep learning in the future.  Meanwhile, the theoretical convergence of networks with delicate structures is also interesting and needs to be analyzed in the future. It will also be interesting to further investigate   problem~\eqref{def_cal} with continuum action space $\mk{L}$, which can be applied to pricing variable annuities problems as recently studied in \cite{GoMoZa:192,GoMoZa:19}. To this end, advanced deep learning algorithms, possibly hybridizing with other numerical techniques to solve partial differential equations for pricing and careful discretization of $\mk{L}$ for convergence, are needed and left for future work.


\section*{Acknowledgment}
We are grateful to Professor Marcel Nutz for bringing the reference \cite{BeChJe:18} into our attention, and Jay Roberts and Kyle Mylonakis for useful discussions.

\bibliographystyle{plain}
\bibliography{Reference}

\begin{thebibliography}{10}

\bibitem{AidLangrene12}
R.~Aid, L.~Campi, N.~Langren{\'e}, and H.~Pham.
\newblock A probabilistic numerical method for optimal multiple switching
  problem and application to investments in electricity generation.
\newblock {\em SIAM Journal of Financial Mathematics}, 5(1):191--231, 2014.

\bibitem{andersen2004primal}
L.~Andersen and M.~Broadie.
\newblock Primal-dual simulation algorithm for pricing multidimensional
  american options.
\newblock {\em Management Science}, 50(9):1222--1234, 2004.

\bibitem{BadrinarayananKendallCipolla17}
V.~Badrinarayanan, A.~Kendall, and R.~Cipolla.
\newblock Seg{N}et: {A} deep convolutional encoder-decoder architecture for
  image segmentation.
\newblock {\em IEEE Trans. Pattern Anal. Mach. Intell.}, 39:2481--2459, 2017.

\bibitem{BeChJe:18}
S.~Becker, P.~Cheridito, and A.~Jentzen.
\newblock Deep optimal stopping, 2018.
\newblock arXiv:1804.05394.

\bibitem{Belomestny11}
D.~Belomestny.
\newblock Pricing bermudan options by nonparametric regression: optimal rates
  of convergence for lower estimates.
\newblock {\em Finance and Stochastics}, 15:655--683, 2011.

\bibitem{Bender11}
C.~Bender.
\newblock Primal and dual pricing of multiple exercise options in continuous
  time.
\newblock {\em SIAM Journal on Financial Mathematics}, 2(1):562--586, 2011.

\bibitem{binois2018replication}
M.~Binois, J.~Huang, R.~B. Gramacy, and M.~Ludkovski.
\newblock Replication or exploration? {S}equential design for stochastic
  simulation experiments.
\newblock {\em Technometrics}, 2018.
\newblock accepted.

\bibitem{bjork2009arbitrage}
Tomas Bj{\"o}rk.
\newblock {\em Arbitrage theory in continuous time}.
\newblock Oxford university press, 2009.

\bibitem{boyle1989numerical}
Phelim~P Boyle, Jeremy Evnine, and Stephen Gibbs.
\newblock Numerical evaluation of multivariate contingent claims.
\newblock {\em The Review of Financial Studies}, 2(2):241--250, 1989.

\bibitem{broadie2008improved}
M.~Broadie and M.~Cao.
\newblock Improved lower and upper bound algorithms for pricing american
  options by simulation.
\newblock {\em Quantitative Finance}, 8(8):845--861, 2008.

\bibitem{BubeckMunos11}
S.~Bubeck, R.~Munos, and G.~Stoltz.
\newblock Pure exploration in finitely-armed and continuous-armed bandits.
\newblock {\em Theoretical Computer Science}, 412(19):1832--1852, 2011.

\bibitem{BubeckMunos11X}
S.~Bubeck, R.~Munos, G.~Stoltz, and C.~Szepesvari.
\newblock X-armed bandits.
\newblock {\em The Journal of Machine Learning Research}, 12:1655--1695, 2011.

\bibitem{Carriere96}
J.~F. Carriere.
\newblock Valuation of the early-exercise price for options using simulations
  and nonparametric regression.
\newblock {\em Insurance: Math. Econom.}, 19:19--30, 1996.

\bibitem{chip:geor:mccu:2010}
H.~A. Chipman, E.~I. George, and R.~E. McCulloch.
\newblock {BART}: {B}ayesian additive regression trees.
\newblock {\em The Annals of Applied Statistics}, 4(1):266--298, 2010.

\bibitem{Keras}
F.~Chollet et~al.
\newblock Keras.
\newblock \url{https://keras.io}, 2015.

\bibitem{Dozat16}
T.~Dozat.
\newblock Incorporating nesterov momentum into adam, 2016.

\bibitem{EHanLi19}
W.~E, J.~Han, and Q.~Li.
\newblock A mean-field optimal control formulation of deep learning.
\newblock {\em Res. Math. Sci.}, 6:10, 2019.

\bibitem{GabillonBubeck11}
V.~Gabillon, M.~Ghavamzadeh, A.~Lazaric, and S.~Bubeck.
\newblock Multi-bandit best arm identification.
\newblock In {\em Advances in Neural Information Processing Systems}, pages
  2222--2230, 2011.

\bibitem{GoMoZa:19}
L.~Gouden{\`e}ge, A.~Molent, and A.~Zanette.
\newblock Gaussian process regression for pricing variable annuities with
  stochastic volatility and interest rate.
\newblock {\em arXiv preprint arXiv:1903.00369}, 2019.

\bibitem{GoMoZa:194}
L.~Gouden{\`e}ge, A.~Molent, and A.~Zanette.
\newblock Machine learning for pricing american options in high-dimensional
  markovian and non-markovian models.
\newblock {\em arXiv preprint arXiv:1905.09474}, 2019.

\bibitem{GoMoZa:192}
L.~Gouden{\`e}ge, A.~Molent, and A.~Zanette.
\newblock Pricing and hedging gmwb in the heston and in the black--scholes with
  stochastic interest rate models.
\newblock {\em Computational Management Science}, 16(1-2):217--248, 2019.

\bibitem{GoMoZa:193}
L.~Gouden{\`e}ge, A.~Molent, and A.~Zanette.
\newblock Variance reduction applied to machine learning for pricing
  bermudan/american options in high dimension.
\newblock {\em arXiv preprint arXiv:1903.11275}, 2019.

\bibitem{gramacy:apley:2013}
R.~B. {Gramacy} and D.~W. {Apley}.
\newblock Local {G}aussian process approximation for large computer
  experiments.
\newblock {\em Journal of Computational and Graphical Statistics},
  24(2):561--578, 2015.

\bibitem{GL13}
R.~B. Gramacy and M.~Ludkovski.
\newblock Sequential design for optimal stopping problems.
\newblock {\em SIAM Journal on Financial Mathematics}, 6(1):748--775, 2015.

\bibitem{GramacyPolson11}
R.~B. Gramacy and N.G. Polson.
\newblock Particle learning of {G}aussian process models for sequential design
  and optimization.
\newblock {\em Journal of Computational and Graphical Statistics},
  20(1):102--118, 2011.

\bibitem{tgpPackage}
R.~B. Gramacy and M.~Taddy.
\newblock tgp, an {R} package for treed {G}aussian process models.
\newblock {\em Journal of Statistical Software}, 33:1--48, 2012.

\bibitem{GTP-trees11}
R.~B. Gramacy, M.~Taddy, and N.~Polson.
\newblock Dynamic trees for learning and design.
\newblock {\em Journal of the American Statistical Association},
  106(493):109--123, 2011.

\bibitem{GrunewalderAudibert10}
S.~Gr{\"u}new{\"a}lder, J.-Y. Audibert, M.~Opper, and J.~Shawe-Taylor.
\newblock Regret bounds for {G}aussian process bandit problems.
\newblock In {\em International Conference on Artificial Intelligence and
  Statistics}, pages 273--280, 2010.

\bibitem{HaRoMyYa:18}
J.~C. Hateley, J.~Roberts, K.~Mylonakis, and X.~Yang.
\newblock Deep learning seismic substructure detection using the frozen
  {G}aussian approximation, 2018.
\newblock arXiv:1810.06610.

\bibitem{HeZhReSu:15}
K.~He, X.~Zhang, S.~Ren, and J.~Sun.
\newblock Deep residual learning for image recognition, 2015.
\newblock CoRR, abs/1512.03385.

\bibitem{hepperger2013pricing}
P.~Hepperger.
\newblock Pricing high-dimensional bermudan options using variance-reduced
  monte-carlo methods.
\newblock {\em Journal of Computational Finance}, 16(3):99--126, 2013.

\bibitem{Ho:91}
K.~Hornik.
\newblock Approximation capabilities of multilayer feedforward networks.
\newblock {\em Neural networks}, 4(2):251--257, 1991.

\bibitem{HuLudkovski17}
R.~Hu and M.~Ludkovski.
\newblock Sequential design for ranking response surfaces.
\newblock {\em SIAM/ASA Journal on Uncertainty Quantification}, 5(1):212--239,
  2017.

\bibitem{jain2015stochastic}
S.~Jain and C.~W. Oosterlee.
\newblock The stochastic grid bundling method: Efficient pricing of bermudan
  options and their greeks.
\newblock {\em Applied Mathematics and Computation}, 269:412--431, 2015.

\bibitem{juneja2009variance}
S.~Juneja and H.~Kalra.
\newblock Variance reduction techniques for pricing american options using
  function approximations.
\newblock {\em Journal of Computational Finance}, 12(3):79, 2009.

\bibitem{Adam}
D.~Kingma and J.~Ba.
\newblock Adam: A method for stochastic optimization.
\newblock {\em arXiv preprint arXiv:1412.6980}, 2014.

\bibitem{kohler2010review}
M.~Kohler.
\newblock A review on regression-based monte carlo methods for pricing american
  options.
\newblock In {\em Recent developments in applied probability and statistics},
  pages 37--58. 2010.

\bibitem{Secomandi11}
G.~Lai, M.~X. Wang, S.~Kekre, A.~Scheller-Wolf, and N.~Secomandi.
\newblock Valuation of storage at a liquefied natural gas terminal.
\newblock {\em Operations Research}, 59(3):602--616, 2011.

\bibitem{LeCunBengioHinton15}
Y.~LeCun and Y.~Bengio.
\newblock Deep learning.
\newblock {\em Nature}, 521:436--444, 2015.

\bibitem{letourneau2014refining}
P.~L{\'e}tourneau and L.~Stentoft.
\newblock Refining the least squares monte carlo method by imposing structure.
\newblock {\em Quantitative Finance}, 14(3):495--507, 2014.

\bibitem{LudkovskiLin14}
J.~Lin and M.~Ludkovski.
\newblock Sequential {B}ayesian inference in hidden {M}arkov stochastic kinetic
  models with application to detection and response to seasonal epidemics.
\newblock {\em Statistics and Computing}, 24(6):1047--1062, 2014.

\bibitem{Longstaff}
F.A. Longstaff and E.S. Schwartz.
\newblock Valuing {A}merican options by simulations: a simple least squares
  approach.
\newblock {\em The Review of Financial Studies}, 14:113--148, 2001.

\bibitem{ludkovski2015kriging}
M.~Ludkovski.
\newblock Kriging metamodels and experimental design for bermudan option
  pricing.
\newblock {\em Journal of Computational Finance}, 22(1):37--77, 2018.

\bibitem{LN10}
M.~Ludkovski and J.~Niemi.
\newblock Optimal dynamic policies for influenza management.
\newblock {\em Statistical Communications in Infectious Diseases}, 2(1):article
  5 (electronic), 2010.

\bibitem{LN11wsc}
M.~Ludkovski and J.~Niemi.
\newblock Optimal disease outbreak decisions using stochastic simulation.
\newblock In {\em Simulation Conference (WSC), Proceedings of the 2011 Winter},
  pages 3844--3853. IEEE, 2011.

\bibitem{lyu2018evaluating}
X.~Lyu, M.~Binois, and M.~Ludkovski.
\newblock Evaluating gaussian process metamodels and sequential designs for
  noisy level set estimation.
\newblock {\em arXiv preprint arXiv:1807.06712}, 2018.

\bibitem{MeinshausenHambly04}
N.~Meinshausen and B.~M. Hambly.
\newblock Monte {C}arlo methods for the valuation of multiple-exercise options.
\newblock {\em Mathematical Finance}, 14(4):557--583, 2004.

\bibitem{MerlGramacy09}
D.~Merl, R.~Johnson, R.~B. Gramacy, and M.~Mangel.
\newblock A statistical framework for the adaptive management of
  epidemiological interventions.
\newblock {\em PLoS ONE}, 4(6):e5087, 2009.

\bibitem{Picheny10}
V.~Picheny, D.~Ginsbourger, O.~Roustant, R.~T. Haftka, and N.-H. Kim.
\newblock Adaptive designs of experiments for accurate approximation of a
  target region.
\newblock {\em Journal of Mechanical Design}, 132:071008, 2010.

\bibitem{RanjanBingham08}
P.~Ranjan, D.~Bingham, and G.~Michailidis.
\newblock Sequential experiment design for contour estimation from complex
  computer codes.
\newblock {\em Technometrics}, 50(4):527--541, 2008.

\bibitem{Adamcvg}
S.~J. Reddi, S.~Kale, and S.~Kumar.
\newblock On the convergence of adam and beyond.
\newblock 2018.

\bibitem{RonnebergerFischerBrox15}
O.~Ronneberger, P.~Fischer, and T.~Brox.
\newblock U-net: Convolutional networks for biomedical image segmentation.
\newblock In {\em MICCAI 2015: Medical Image Computing and Computer-Assisted
  Intervention}, pages 234--241, 2015.

\bibitem{kmPackage-R}
Olivier Roustant, David Ginsbourger, and Yves Deville.
\newblock Dicekriging, {DiceOptim}: Two {R} packages for the analysis of
  computer experiments by kriging-based metamodeling and optimization.
\newblock {\em Journal of Statistical Software}, 51(1):1--51, 2012.

\bibitem{ShelhamerLongDarrell15}
E.~Shelhamer, J.~Long, and T.~Darrell.
\newblock Fully convolutional networks for semantic segmentation.
\newblock In {\em 2015 IEEE Conference on Computer Vision and Pattern
  Recognition (CVPR)}, pages 3431--3440, 2015.

\bibitem{SpDoBrRi:14}
J.~T. Springenberg, A.~Dosovitskiy, T.~Brox, and M.~A. Riedmiller.
\newblock Striving for simplicity: {T}he all convolutional net, 2014.
\newblock CoRR, abs/1412.6806.

\bibitem{tsitsiklis2001regression}
J.~N. Tsitsiklis and B.~Van~Roy.
\newblock Regression methods for pricing complex american-style options.
\newblock {\em IEEE Transactions on Neural Networks}, 12(4):694--703, 2001.

\end{thebibliography}

\end{document}